\documentclass[journal]{IEEEtran}

\usepackage{cite}
\usepackage{graphicx}
\usepackage{balance}  

\usepackage{amsfonts,amssymb,amsmath,bm}
\usepackage{makecell}
\usepackage[caption=false,font=normalsize,labelfont=sf,textfont=sf]{subfig}
\usepackage{multirow}
\usepackage{verbatim}
\usepackage{booktabs}
\usepackage[shortlabels]{enumitem}
\usepackage{wasysym}
\usepackage{algorithmic}
\usepackage{arydshln}
\usepackage{chngcntr} 
\usepackage{xcolor}
\usepackage{xspace}
\usepackage[linesnumbered,vlined,ruled]{algorithm2e}

\usepackage{color}

\newcommand{\lh}[1]{\textcolor{black}{#1}}

\newcommand{\method}{\textsf{MuseGraph}\xspace}

\usepackage{amsthm}

\newcounter{example}[section]

\def\BibTeX{{\rm B\kern-.05em{\sc i\kern-.025em b}\kern-.08em
    T\kern-.1667em\lower.7ex\hbox{E}\kern-.125emX}}

\usepackage[utf8]{inputenc} 
\usepackage[T1]{fontenc}    
\usepackage{hyperref}       
\usepackage{url}            
\usepackage{nicefrac}       
\usepackage{microtype}      

\begin{document}


\title{\raisebox{-0.6ex}{\includegraphics[height=1cm]{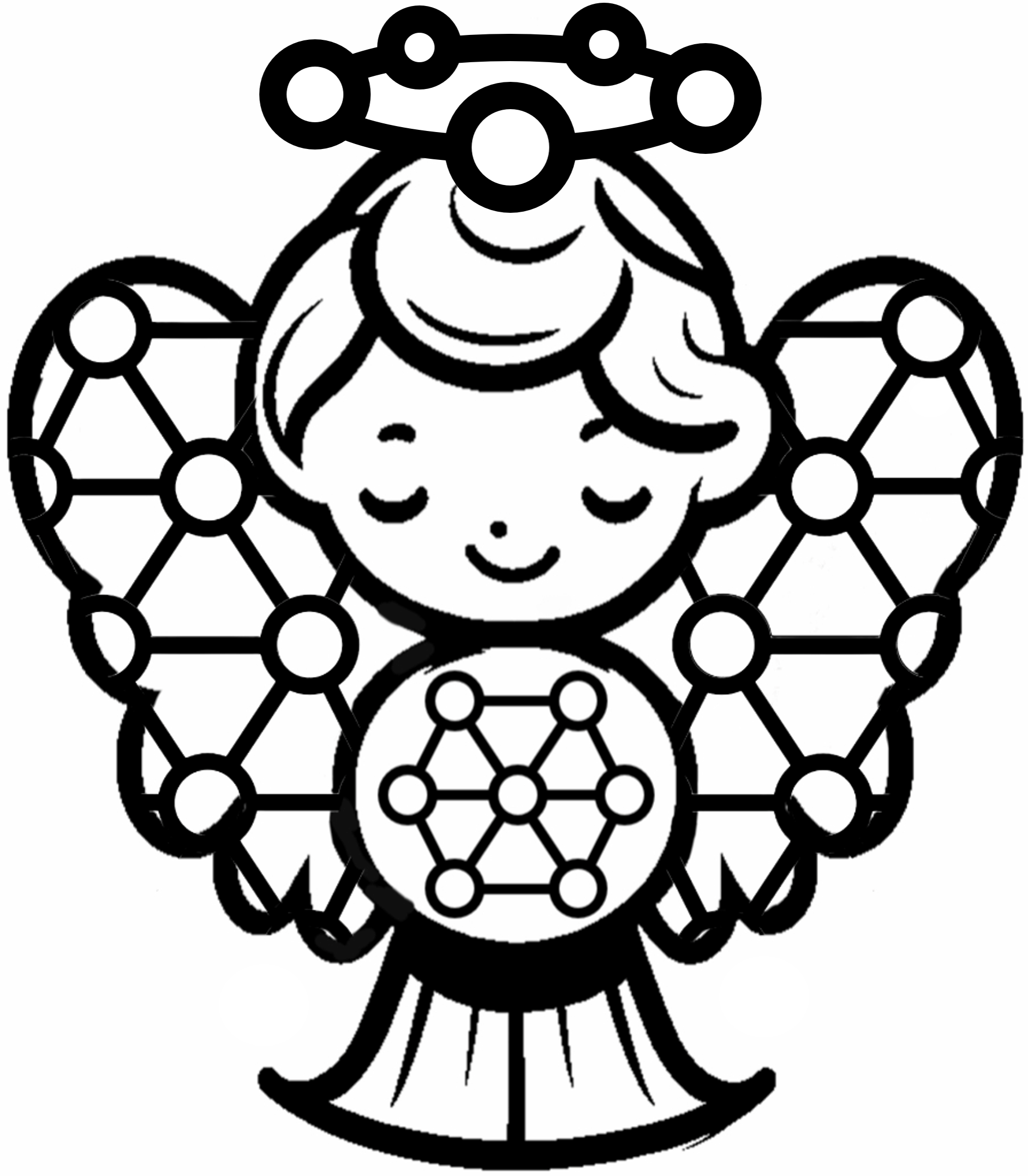}} Graph-oriented Instruction Tuning of Large Language Models for Generic Graph Mining}


%
\author{Yanchao~Tan~\IEEEmembership{Member,~IEEE},
        Hang~Lv,
        Pengxiang~Zhan,
        Shiping~Wang~\IEEEmembership{Senior Member,~IEEE},
        Carl~Yang~\IEEEmembership{Member,~IEEE}
        
        \IEEEcompsocitemizethanks{
        \IEEEcompsocthanksitem Y. Tan, H. Lv, P. Zhan, and S. Wang are with the Engineering Research Center of Big Data Intelligence, Ministry of Education; the Fujian Key Laboratory of Network Computing and Intelligent Information Processing; and the College of Computer and Data Science, Fuzhou University, Fuzhou 350116, China, Email: yctan@fzu.edu.cn, lvhangkenn@gmail.com, yyyzhanpengxiang@163.com, shipingwangphd@163.com.
        \IEEEcompsocthanksitem C. Yang (Corresponding Author) is with the Department of Computer Science, Emory University, Atlanta 30322, United States, E-mail: j.carlyang@emory.edu.
        }
}

\IEEEtitleabstractindextext{
\begin{abstract}
Graphs with abundant attributes are essential in modeling interconnected entities and enhancing predictions across various real-world applications.
Traditional Graph Neural Networks (GNNs) often require re-training for different graph tasks and datasets.
Although the emergence of Large Language Models (LLMs) has introduced new paradigms in natural language processing, their potential for generic graph mining—training a single model to simultaneously handle diverse tasks and datasets—remains under-explored.
To this end, our novel framework \method, seamlessly integrates the strengths of GNNs and LLMs into one foundation model for graph mining across tasks and datasets.
This framework first features a compact graph description to encapsulate key graph information within language token limitations.
Then, we propose a diverse instruction generation mechanism with Chain-of-Thought (CoT)-based instruction packages to distill the reasoning capabilities from advanced LLMs like GPT-4.
Finally, we design a graph-aware instruction tuning strategy to facilitate mutual enhancement across multiple tasks and datasets while preventing catastrophic forgetting of LLMs' generative abilities.
Our experimental results demonstrate significant improvements in five graph tasks and ten datasets, showcasing the potential of our \method in enhancing the accuracy of graph-oriented downstream tasks while improving the generation abilities of LLMs.
\end{abstract}

}

\markboth{IEEE Transactions on Pattern Analysis and Machine Intelligence,~Vol.~47, No.~8, August~2025}
{Graph-oriented Instruction Tuning of Large Language Models for Generic Graph Mining}

\maketitle
\IEEEdisplaynontitleabstractindextext

\IEEEpeerreviewmaketitle


\section{Introduction}
\label{sec:intro}

\IEEEPARstart{G}{raphs} with plentiful attributes are widely used to model interconnected real-world entities, and they are pivotal for improving downstream predictions across diverse real-world applications. 
Recently, Graph Neural Networks (GNNs) have been commonly adopted for modeling attributed graphs~\cite{hamilton2017inductive,cui2020adaptive}.
However, they are usually trained on specific tasks and datasets and need to be re-trained whenever applied to different ones.
Inspired by the great success of Large Language Models (LLMs), the combination of GNNs and LLMs aims to enhance the processing of text-attributed graphs, which can improve the model’s capabilities across various tasks and datasets.
Existing studies can be categorized into two main approaches.
The first category tries to train GNNs with LLM-enhanced features (e.g., LLM-GNN~\cite{chen2023label}, TAPE~\cite{he2023harnessing}, OFA~\cite{liu2023one}, ALL-in-One~\cite{sun2023all}, and GHGRL~\cite{gao2025bootstrapping}).
The second category explores LLMs for various graph applications (e.g., GPT4Graph~\cite{guo2023gpt4graph}, GraphGPT~\cite{tang2024graphgpt}, 
HiGPT~\cite{tang2024higpt}, NLGraph~\cite{wang2024language}, and InstructGLM~\cite{ye2023natural}).

\begin{figure}
    \centering
    \includegraphics[width=1\linewidth]{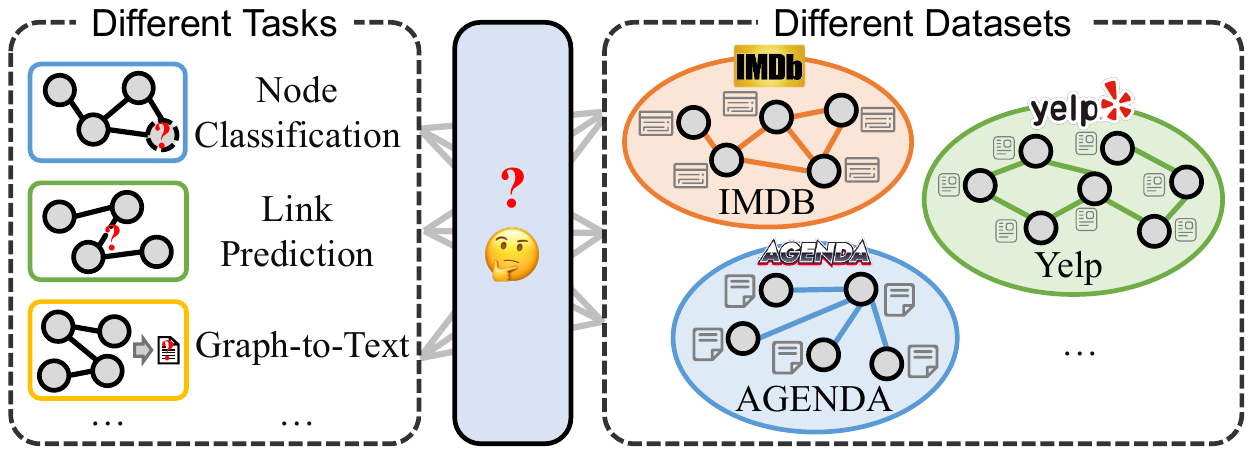}
     \caption{An illustrative toy example of the need for a generic graph model that can be directly applied to various graph-related tasks and datasets.}
    \label{fig:toy}
    \vspace{-0.4cm}
\end{figure}

Despite the promising direction of integrating LLMs with GNNs, the full exploration of LLMs' power and generation capabilities for graphs has yet to be under-explored~\cite{sun2023all,liu2023one,huang2024prodigy}, where one powerful graph model can be trained on multiple tasks/datasets and generalize well to more tasks/datasets.
Fig.~\ref{fig:toy} demonstrates the necessity for a generic graph framework, showcasing a wide range of task and dataset combinations.
Such a generic graph model can capture the semantic and structural information not only for various graph tasks (e.g., node classification, link prediction, and graph-to-text) but also for diverse graph datasets (e.g., from IMDB's movie data to AGENDA's academic texts). 
This enables one single foundation model to understand graph data by generating contextually rich textual interpretations that are central to LLMs.
However, three key challenges hinder the achievement of this goal.

\noindent
\textbf{Challenge \uppercase\expandafter{\romannumeral1}:} \emph{How to extract informative graph descriptions under the limitation of language tokens?} 
To harness the full potential of LLMs for generic graph mining, an essential obstacle is translating the graph with abundant semantics and complex structures into a format that LLMs can process effectively, especially under strict language token limitations.
Without a compact graph description that accurately encapsulates key information from graphs within the LLMs' token limitations, the model's capabilities to grasp and utilize the graph's semantic and structural richness is severely limited, potentially leading to suboptimal performance in graph applications.

\noindent
\textbf{Challenge \uppercase\expandafter{\romannumeral2}:} \emph{How to automatically generate diverse instructions?}
Creating a diverse set of high-quality instructions for fine-tuning LLMs is fundamental for generic graph tasks.
However, these instructions are often difficult for LLMs to comprehend when encountering unfamiliar graph structures or new tasks, and are costly to produce manually.
While advanced LLMs like GPT-4 possess the capabilities to understand and reason diverse instructions, it remains unknown how to effectively and efficiently leverage the reasoning abilities of advanced LLMs to produce relevant and task-specific instructions.

\noindent
\textbf{Challenge \uppercase\expandafter{\romannumeral3}:} \emph{How to properly allocate instructions for graph-oriented instruction tuning?}
The effectiveness of instruction tuning for LLMs largely depends on how the instructions are structured, enabling LLMs to accurately understand and execute graph-related tasks. 
Balancing a variety of tasks and datasets presents a significant challenge in facilitating mutual enhancement across these graph applications while preventing catastrophic forgetting of LLMs' generative abilities.

To tackle these challenges, we propose Graph-oriented Instruction Tuning of Large Language Models for Generic Graph Mining (\method), which consists of three pivotal steps: (i) \emph{Development of Compact Graph Descriptions}, where we introduce a novel ``node energy'' metric to textualize graphs with essential semantic and structural details under limited language tokens; (ii) \emph{Generation of Diverse Instructions}, which distills the reasoning abilities of advanced LLMs like GPT-4 to create Chain-of-Thought (CoT)-based instruction packages tailored for various graph tasks, thus enriching LLMs' capabilities in understanding and analyzing graph data without the expense of manual instruction crafting; (iii) \emph{Graph-aware Instruction Tuning}, which introduces a dynamic instruction package allocation strategy based on the specific needs of each graph task, ensuring comprehensive and effective LLM tuning.

Our overall contributions are summarized as follows:
\begin{itemize}[leftmargin=10pt]
\item \emph{Formulation of generic graph mining. }We establish a generic graph framework that effectively and efficiently transforms graph semantics and structures into LLM-friendly formats while enhancing the generation capabilities necessary for diverse graph-related tasks.
\item \emph{Effective model designs.}
We design and implement a set of models and mechanisms, including the development of compact graph descriptions, automatically generating diverse task-specific Chain-of-Thought (CoT)-based instruction packages, and graph-aware instruction tuning, targeting a unified graph model across tasks and datasets.
\item \emph{Extensive experiments across graph tasks and datasets.} We conduct thorough experiments to validate our approach with five tasks and ten datasets, demonstrating its superiority over existing state-of-the-art methods and highlighting its effective, generative, and interpretable abilities in enhancing generic graph mining. 
\end{itemize}

\section{Related Work}

\subsection{Semantic-rich Graph Representation Learning}
Graph representation learning has emerged as a key technique for the complex structures of networks with abundant attributes~\cite{yang2020heterogeneous,cui2020adaptive,DBLP:journals/tkde/CaiZC18,mao2023hinormer}. 
Many existing node embedding approaches have explored and harnessed the significant potential of Random Walks (RWs) in preserving the graph topological structures~\cite{grover2016node2vec,dong2017metapath2vec,ivanov2018anonymous,li2021higher}.
However, these approaches overlook the rich attribute information of nodes and edges on the graph~\cite{mao2023hinormer}.
Recently, Graph Neural Networks (GNNs) learn node representations through aggregating content information from neighbor nodes while preserving the surrounding structure of graphs~\cite{kipfsemi,hamilton2017inductive,yang2020relation}.
However, most current GNNs are trained within a supervised learning setting, which demands a large amount of task-specific labeled data and may not always be available in real-world scenarios~\cite{li2023towards,qian2020attribute}.
Moreover, the learned embeddings often lack adaptability across different downstream tasks~\cite{li2021higher}, which have to be re-trained whenever applied to various graph applications.

Despite the efforts to reduce reliance on labeled data through pre-training expressive GNNs through self-supervised methods (e.g., contrastive learning~\cite{zhang2023multi,li2023towards,jiang2021pre}), their effectiveness in specific downstream tasks still significantly depends on the appropriate choice of suitable self-supervision tasks and attribute encoders~\cite{liu2023hierarchical,sun2022gppt}.
Therefore, there is still a lack of a uniform framework for generic graph mining across different tasks and datasets.

\subsection{Leveraging LLMs for Graph Mining}

 \begin{figure*}[ht]
 \setlength{\abovecaptionskip}{0.1cm}
    \centering
\includegraphics[width=1\linewidth]{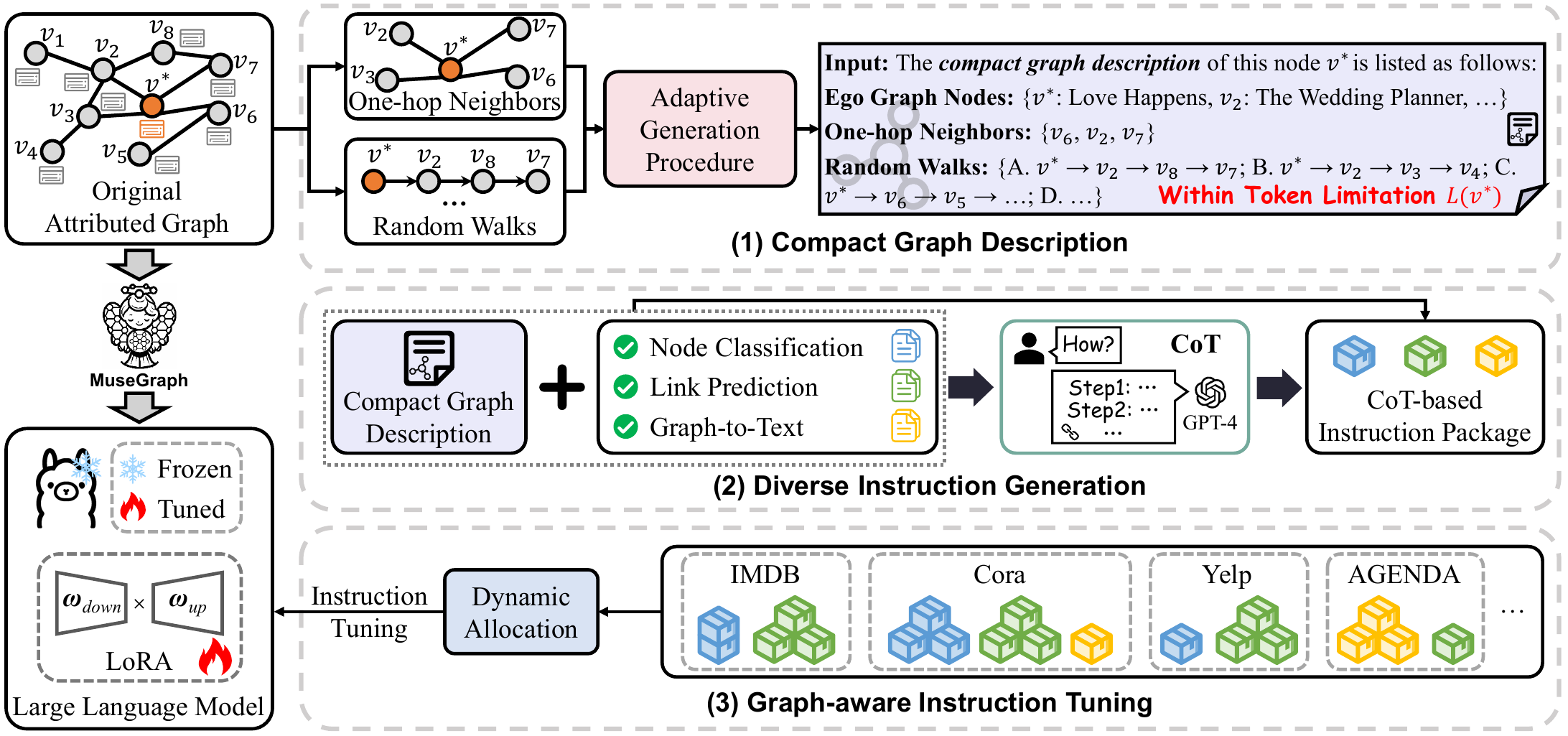}
    \caption{The overall of \method, which consists of Compact Graph Description, Diverse Instruction Generation, and Graph-aware Instruction Tuning.}
    \label{fig:framework}
\end{figure*}

\begin{table}
\setlength{\abovecaptionskip}{0.1cm}
  \centering
  \caption{A comparison between \method and related methods, \lh{where ``GNN'' and ``LLM'' refer to methods that train GNNs and LLMs as predictors, respectively.}}
  \begin{tabular}{p{2.3cm}cc|cc}
    \toprule
    & \lh{GNN} & \lh{LLM} & Cross-Task & Cross-Dataset \\
    \midrule
    LLM-GNN~\cite{chen2023label} & \CIRCLE & \Circle & \Circle & \Circle \\ 
    TAPE~\cite{he2023harnessing} & \CIRCLE & \Circle & \Circle & \Circle \\
    OFA~\cite{liu2023one} & \CIRCLE & \Circle & \CIRCLE & \CIRCLE \\
    All-in-One~\cite{sun2023all} & \CIRCLE & \Circle & \CIRCLE & \CIRCLE \\
    GHGRL~\cite{gao2025bootstrapping} & \CIRCLE & \Circle & \Circle & \Circle \\
    
    GPT4Graph~\cite{guo2023gpt4graph} &\Circle  & \Circle  & \Circle  & \Circle \\
    GraphGPT~\cite{tang2024graphgpt} & \Circle & \CIRCLE & \Circle & \CIRCLE \\
    HiGPT~\cite{tang2024higpt} & \Circle & \CIRCLE  & \Circle  & \CIRCLE \\
    NLGraph~\cite{wang2024language}  & \Circle  & \Circle  & \Circle & \Circle \\
    InstructGLM~\cite{ye2023natural} & \Circle & \CIRCLE & \Circle & \Circle \\
    \hline
    \method & \Circle & \CIRCLE & \CIRCLE & \CIRCLE \\
    \bottomrule
  \end{tabular}
  \label{tab:comparison}
\end{table}

Inspired by the remarkable advancements in Large Language Models (LLMs), integrating Graph Neural Networks (GNNs) with LLMs is creating substantial progress in handling complex text-attributed graphs~\cite{li2024survey,chen2023label}. 
Existing approaches that leverage this integration can be broadly divided into the following two categories (shown in Table~\ref{tab:comparison}).

\lh{The first category is training GNNs as predictors, augmented with LLM-enhanced features, labeled ``GNN'' in Table~\ref{tab:comparison}.} 
For example, the model LLM-GNN~\cite{chen2023label} used LLMs as annotators to generate the pseudo labels of nodes used to train GNNs.
TAPE~\cite{he2023harnessing} prompted LLMs to obtain the explanations of predictions to enhance node initial features.
PRODIGY~\cite{huang2024prodigy} utilized LLMs to encode the textual information associated with nodes on the graph.
OFA~\cite{liu2023one} described all nodes and edges using human-understandable prompt texts and converted them into embedded features by LLMs.
All-in-One~\cite{sun2023all} reformulated different graph tasks with the unified graph prompts and language prompts.
\lh{GHGRL~\cite{gao2025bootstrapping} employed LLMs to automatically summarize and classify various data formats and types, aligning node features.}
These approaches primarily leveraged LLMs to process textual content within graphs and are often optimized for specific domains.
However, they generally struggle with cross-task or cross-dataset applications, limiting the generation capabilities of LLMs across various graph tasks and datasets.

\lh{The second category, denoted as ``LLM'' in Table~\ref{tab:comparison}, mainly focuses on utilizing LLMs to directly perform graph tasks.}
For example, \lh{GPT4Graph~\cite{guo2023gpt4graph} and NLGraph~\cite{wang2024language} proposed graph-based benchmarks,} evaluating the understanding and analytical capabilities of LLMs on graph data with designed natural language prompts.
\lh{Notably, although they do not fine-tune LLMs on graph structures, their benchmarks effectively uncover current limitations and future directions for improving LLMs' capabilities in graph comprehension and reasoning.}
Both GraphGPT~\cite{tang2024graphgpt} and HiGPT~\cite{tang2024higpt} effectively combined task-specific GNNs with one projector on top of LLMs, enhancing performance across different datasets.
InstructGLM~\cite{ye2023natural} tuned LLMs with scalable graph prompts based on natural language instructions, where it can integrate node features from trained GNNs to improve the accuracy of node classification tasks.
Compared with GNN-based methods, LLM-based methods have the potential to perform generic graph mining with powerful cross-task and cross-dataset abilities.
\lh{However, such potential remains largely under-explored, as many existing LLM-based approaches still rely on node representations generated by GNNs tailored to specific tasks/datasets.}
According to the characteristics of different models (shown in Table~\ref{tab:comparison}), there is a pressing need for a generic graph framework that can accurately understand graph data while enhancing generation capabilities across diverse tasks and datasets.

\section{The \method Framework}
\label{sec:method}

\subsection{Overview of Our Framework}
\noindent\textbf{Objective:}
In this paper, we aim to develop a unified framework that can seamlessly integrate the strengths of Graph Neural Networks (GNNs) and Large Language Models (LLMs) into one foundation model via graph-oriented instruction tuning of LLMs.
Through this, we seek to enable a more effective and generic approach for graph mining across various downstream tasks and datasets.

\noindent\textbf{Overview:}
To achieve this goal, we propose \method framework, which comprises three major components.
Firstly, we develop a compact graph description mechanism that captures critical semantic and structural details within the constraints of language token limitations.
Secondly, we generate a diverse range of instructions via the reasoning capabilities of advanced LLMs like GPT-4, thus facilitating task-specific Chain-of-Thought (CoT)-based instruction packages for graph tasks.
Thirdly, we adopt a graph-aware instruction tuning, utilizing a dynamic allocation strategy for instruction packages tailored to the unique requirements of each graph task and dataset.
The overall model architecture is shown in Fig.~\ref{fig:framework}, and we elaborate on the three main components.

\subsection{Compact Graph Description}
\label{sub:H}
Leveraging the capabilities of LLMs for graphs presents a unique set of challenges, primarily due to LLMs' inherent limitations in directly processing textual interpretations within language token constraints. 
Therefore, it is non-trivial to automatically encapsulate key information from graphs under the token limitations, which requires a compact description including complex node and edge attributes along with structural details to accurately describe the graph. 

Inspired by common graph analysis techniques such as neighbors and walks, we propose a novel method of textualization to describe graphs via these concepts. In this way, neighbors are helpful to understand local connectivity and feature distribution~\cite{kipfsemi,hamilton2017inductive,yang2020relation}, providing a granular view of node attributes; while walks offer a dynamic method to explore the graph's structure and the high-order relationships between nodes, highlighting the diversity of connectivity and paths~\cite{li2021higher,ivanov2018anonymous}. The integration of neighbors and walks can achieve a holistic understanding of graph structures.
\lh{Notably, this textualization method integrates GNN-inspired structural priors (i.e., the core principles of message passing and high-order propagation) into the graph description design, without requiring training specific GNNs.
This enables more powerful and adaptive capabilities across diverse tasks and datasets.}

Given LLM token limitations and the varied contributions of neighbors and walks to node understanding, we further develop an adaptive generation procedure to ensure the compactness of the description. Specifically, we first design a ``node energy''  metric $H(\cdot)$, assessing node information from two perspectives: token count in node attributes and node degree count.
This metric enables us to effectively filter and select neighbor nodes and walk nodes, prioritizing those that are abundant in semantic information and possess a significant number of neighbor nodes, thus enhancing the expressiveness of the graph description.
The calculation of node energy $H(v^*)$ is formulated as follows:
\begin{equation}
    H(v^*) = T(v^*) \times \lceil\log(D(v^*)+1)\rceil,
\label{eq:h}
\end{equation}
where $v^*$ is the target node, ${T(v^*)}$ is the number of node tokens processed by a language tokenizer, ${D(v^*)}$ is the number of node degrees, and $\lceil\cdot\rceil$ is the ceiling operator.

\begin{algorithm}[!t]
\renewcommand{\algorithmicrequire}{\textbf{Input:}}
\renewcommand{\algorithmicensure}{\textbf{Output:}}
\caption{Procedure for generating the compact graph description tailored to a specified node. }
\label{algorithm_simplified}
\begin{algorithmic}[1]
\REQUIRE{Attributed graph $\mathcal{G}$ with $N$ nodes, token count \\set $\mathcal{T}$, node energy set $\mathcal{H}$, target node $v^*$, token limitation $L(v^*)$}
\ENSURE{Key neighbor set $\mathcal{N}(v^*)$, key walk set $\mathcal{W}(v^*)$}
\STATE Initialize $\mathcal{N}(v^*)$, $\mathcal{W}(v^*)$ as empty sets;
\STATE Select $v_i \in G(v^*)$ with $H(v_i) \geq H(v^*)$ and $L(v^*) \geq T(v_i)$ constraints for $\mathcal{N}(v^*)$, where $G(v^*)$ \\is $v^*$'s one-hop neighbors and $H(\cdot)$ can be \\calculated according to Eq.~\ref{eq:h};
\STATE Expand $\mathcal{W}(v^*)$ starting from $v^*$ based on $\mathcal{G}$ within \\$L(v^*)$ and $H(v^*)$ constraints.
\end{algorithmic}
\end{algorithm}

\begin{figure}
    \centering
    \includegraphics[width=1\linewidth]{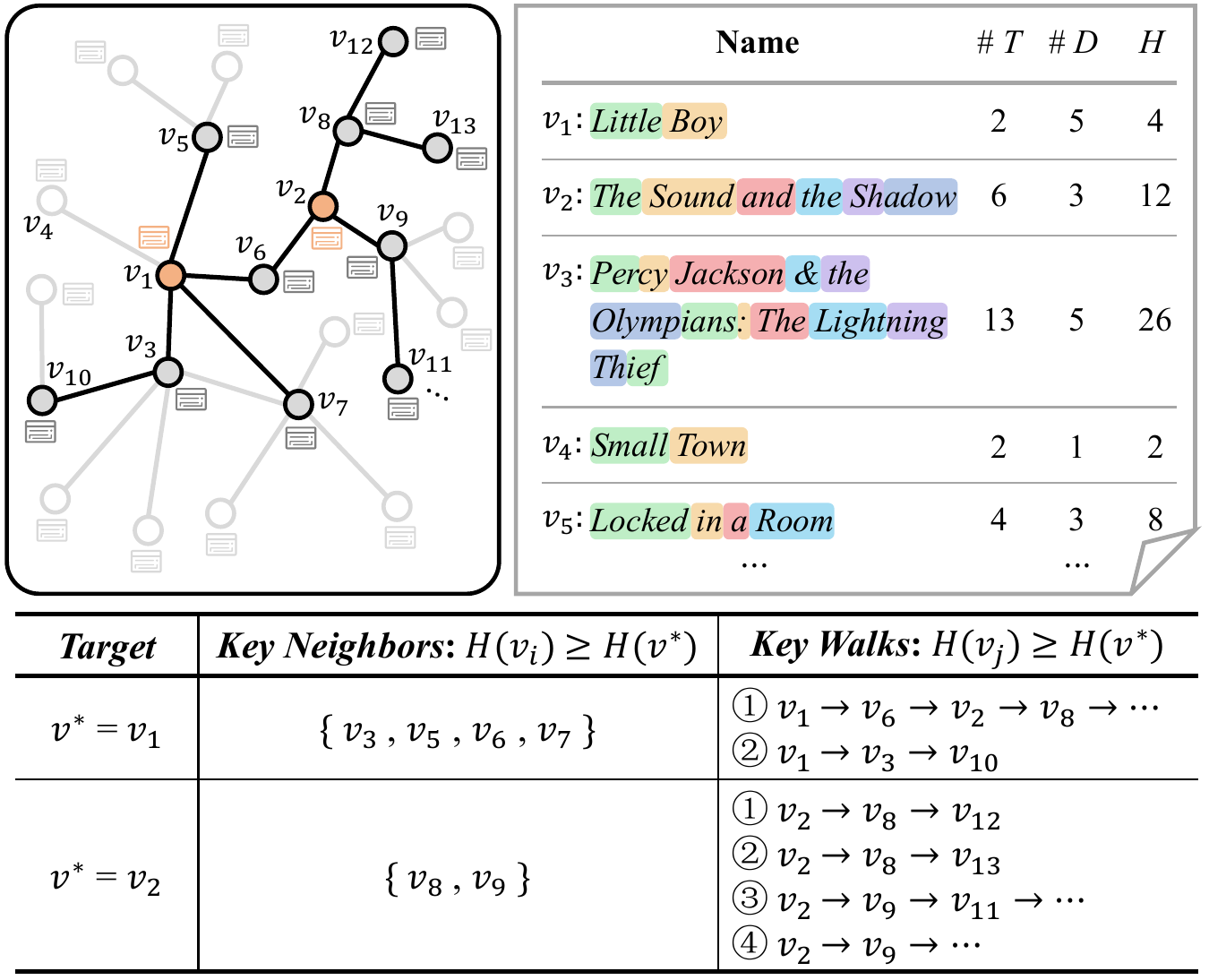}
     \caption{Two examples on IMDB illustrate how to extract the key information of nodes via neighbors and walks based on node energy $H(v^*)$ and the language token limitation $L(v^*)$.}
    \label{fig:h}
\end{figure}

Based on $v^*$, our method strategically incorporates neighbors and walks, as depicted in Algorithm~\ref{algorithm_simplified}. A neighbor $v_i$ is chosen to describe the target $v^*$ if its $H(v_i)$ surpasses $H(v^*)$, ensuring the included node can provide supplemental information. 
The process of adding walks concludes when encountering a node whose $H(v_i)$ does not meet the threshold set by the target's $H(v^*)$, thus refining the input to maximize related graph information within the constraints of the token limit.
To further demonstrate the adaptive input generation for neighbors and walks tailored to each node, we present an example in Fig.~\ref{fig:h}.
Node $v_1$'s with a relatively low $H(v_1)$ value includes a wide range of neighbors to capture more context, excluding $v_4$ due to its even lower $H(v_4)$. Given the token limitation $L(v_1)$, only two walks are sampled for $v_1$ to maintain a compact description.
Conversely, $v_2$ with its higher $H(v_2)$, inherently carries more information, prompting the selection of fewer neighbors. $v_6$ is excluded since $H(v_6)< H(v_2)$. This frees up tokens to detail more walks for $v_2$. The $H$ metric thus effectively balances neighbor and walk inclusion for each node, marrying information-rich and token-efficient characteristics.
The process culminates in the textualization of each node's key information, producing a tailored and compact graph description as depicted in the upper right of Fig.~\ref{fig:framework}.
Note that, a graph-related task can involve multiple nodes (such as link prediction). In this case, we adaptively allocate the token limit requirement of each node involved by calculating the ratio of all node energies using the softmax function.
Additionally, we provide the ablation in Section~\ref{sec:ablation_compact_graph} to further verify the effectiveness of the compact graph description.

\noindent\lh{\textbf{Remark.} Given an attributed graph $\mathcal{G} = (\mathcal{V}, \mathcal{E})$ and an LLM token limitation $L$, the expected utility of including a node $v \in \mathcal{V}$ in the compact graph description can be approximated by the product of its semantic complexity $T(v)$ and structural centrality $D(v)$ as follows:
\begin{equation}
\begin{aligned}
\mathbb{E}[\phi(v)] \propto T(v)& \times \lceil\log(D(v)+1)\rceil = H(v),\\
\text{s.t.} &\sum_{v \in G} T(v) \leq L,
\end{aligned}
\end{equation}
where $\phi(v)$ denotes the information gain of including a node $v$. Under the LLM token constraint $\sum_{v \in G} T(v) \leq L$, prioritizing nodes by descending $H(v)$ value leads to a compact yet informative subgraph $G$, which captures both semantic richness and structural connectivity of the original graph $\mathcal{G}$. Notably, this formulation models node selection as a constrained expected utility maximization problem~\cite{fazel2005network}, where $H(v)$ acts as a practical proxy for the expected informativeness per node. Empirically, longer token spans tend to encode richer semantics~\cite{lian2025lbpe,acevedo2025approach}, while high-degree nodes often serve as key hubs that support global connectivity~\cite{melton2023muxgnn,wangtopological,gao2025graph,gao2023addressing}. Therefore, the node energy $H(v)$ naturally balances token cost with information gain, making it a suitable metric for adaptive neighbor and walk selection within limited LLM tokens.}

\subsection{Diverse Instruction Generation}
With the compact graph descriptions tailored to each node, another critical step in fine-tuning LLMs for generic graph mining is crafting diverse and high-quality instructions~\cite{ouyang2022training,wang2023far}.
Benefiting from the key graph information captured by our designed compact graph descriptions, we can accurately and efficiently construct abundant graph-related task instructions (e.g., node classification, link prediction, and graph-to-text).
While these instructions enable LLMs to effectively grasp compact graph descriptions, they may not always be easily comprehended when encountering unfamiliar graph structures or new tasks, potentially leading to inaccuracies and limited reasoning across various datasets.
Moreover, manual construction of these instructions is often highly costly in terms of both time and resources.

To address these problems, we propose to distill the reasoning capabilities from advanced LLMs~\cite{shridhar2023distilling} (e.g., GPT-4~\cite{achiam2023gpt} with over 200 billion parameters) for graph-related tasks. 
Our approach, inspired by the Chain-of-Thought (CoT) processing methodology~\cite{wei2022chain}, prompts GPT-4 via a flexible template based on our compact graph descriptions for different tasks and then constructs task-specific CoT-based instruction packages.
Different from the existing methods that leverage CoT in the prompting stage~\cite{wei2022chain,xiang2024badchain}, we directly construct CoT-based instruction packages, harnessing their diversity and reasoning abilities to facilitate instruction tuning with graph data.

\begin{figure}[tb]
    \centering
    \includegraphics[width=1\linewidth]{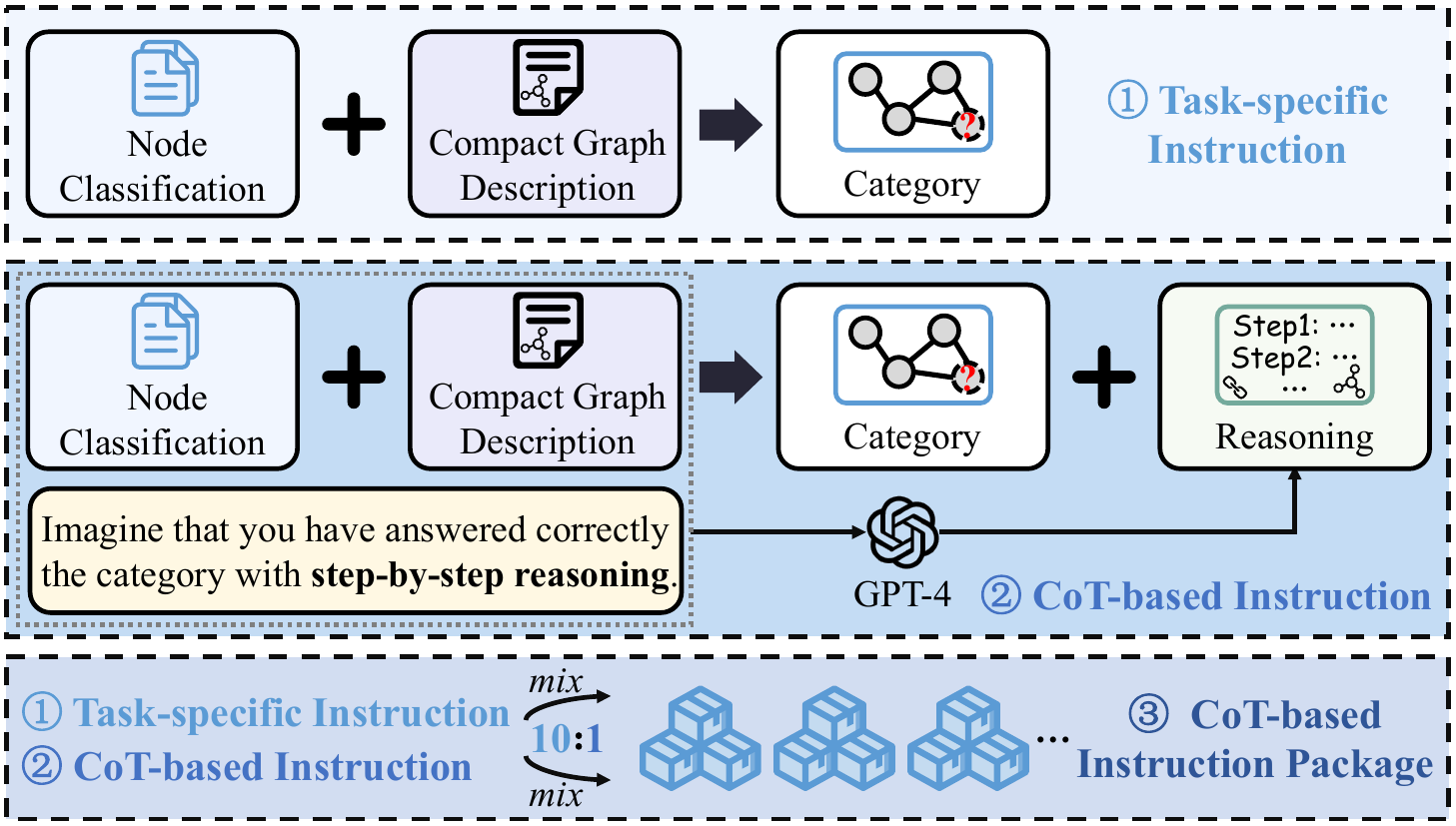}
     \caption{A process showing how to conduct the Chain-of-Thought (CoT)-based instruction package for node classification. We leverage the reasoning ability distilled from advanced LLMs (e.g., GPT-4) and integrate them with task-specific instructions via a 1:10 mix ratio.}
    \label{fig:COT}
\end{figure}

Specifically, as shown in Fig.~\ref{fig:COT}, we first design diverse task-specific instructions based on compact graph descriptions, such as node classification, link prediction, and graph-to-text.
Then, we prompt GPT-4 with a small number of task-specific instructions to generate the corresponding step-by-step reasoning, which is used to conduct CoT-based instructions with the initial instructions across diverse tasks.
This approach distills GPT-4's vast knowledge base to augment the reasoning and analytical abilities of our \method, enhancing the understanding of compact graphs.

To optimize the cost-effectiveness of querying GPT-4, we introduce the CoT-based instruction package. 
For every set of 1,000 standard task-specific instructions, we integrate 100 CoT-based instructions tailored to the same graph task.
This approach not only proves to be economical but also broadens the diversity and flexibility of instructions, accommodating a range of graph-related tasks.

\subsection{Graph-aware Instruction Tuning}
With the proposed compact graph descriptions and diverse instruction generation mechanisms, identifying an effective instruction tuning method for LLMs remains crucial. Such tuning enables LLMs to understand and perform well a wide range of graph mining tasks across multiple datasets and contexts~\cite{wang2023far,wang2022self}.
However, a persistent challenge is the catastrophic forgetting issue commonly faced by LLMs~\cite{mosbach2020stability,vijay2022nerda}. This phenomenon complicates the capabilities to maintain extensive task and dataset coverage without compromising previously acquired knowledge, necessitating strategies that balance new learning with memory retention.

To this end, we propose a dynamic instruction package allocation strategy to adaptively adjust the volume of task-specific CoT-based instruction packages based on the complexities of tasks and datasets, which ensures that more complex tasks/datasets receive a proportionally larger set of instructions for detailed guidance.
The calculation process includes two aspects as follows:
\begin{itemize}
\item For {task complexity}: We assess the complexity of tasks by calculating the average number of answer tokens for each task, which helps in tailoring the CoT-based instruction packages to specific needs within a dataset. \lh{A larger average number of answer tokens indicates a higher level of reasoning and response complexity, enabling us to proportionally allocate more instructions to such tasks.}
\item For {dataset complexity}: We calculate the total node energy $H(\cdot)$ (cf., Eq.~\ref{eq:h}) for each graph data, utilizing this metric to optimize the instruction distribution.
\lh{A higher overall node energy reflects richer semantic and structural characteristics within a dataset, guiding the allocation of more instructions to these information-dense graphs.}
\end{itemize}
By performing this dynamic allocation of instructions, we enhance LLMs' abilities to learn and retain extensive knowledge across a diverse range of graph mining challenges. 
\lh{We provide the ablation in Section~\ref{sec:ablation_dynamic_allocation strategies} to further assess the effectiveness of the dynamic instruction package allocation strategy.}

To balance the effectiveness with computational efficiency, we adopt a graph-aware instruction tuning mechanism, which can sufficiently utilize diverse and high-quality instructions for fine-tuning LLMs~\cite{ouyang2022training}.
Specifically, we adopt a general LLM LLaMA3-8B~\cite{dubey2024llama} with LoRA~\cite{hu2021lora} as our initial fine-tuning point by default.
Then, based on the diverse instruction package set $\mathcal{I}=\{I_1, I_2,\dots, I_D\}$ as model input, we adopt the negative log-likelihood loss as the fine-tuning objective as follows:
\begin{equation}
p_{\theta}\left(Y_{j,k}|I_j,Y_{j,<k}\right)=LLM_{\theta}\left(I_j,Y_{j,<k}\right),
\label{eq:LLM_p}
\end{equation}
\begin{equation}
\mathcal{L}_{\theta}=-\sum_{k=1}^{|Y_j|}{\log p_{\theta}\left(Y_{j,k}|I_j,Y_{j,<k}\right)},
\label{eq:FT}
\end{equation}
where $\theta$ is the learnable parameters of our proposed graph-aware LLM (i.e., \method), $I_j \in \mathcal{I}$ is the input of LLM, and $Y_j$ is the output of LLM.

After obtaining the fine-tuned LLM for generic graph mining, we can apply it to various downstream tasks, such as node classification, link prediction, and graph-to-text.
Notably, our \method can achieve superior performance across diverse tasks and datasets with few instruction packages based on the graph-aware instruction tuning mechanism, even in the few-shot and zero-shot settings (shown in Table~\ref{tab:exp-few-shot-gt} and Fig.~\ref{fig:dyval-bar}).

\noindent\textbf{Model Extension.} Our proposed \method establishes a generic approach for integrating LLMs with graph data.
We can flexibly incorporate a variety of foundation LLMs, as discussed in Section~\ref{exp:foundation_model}, which enhances its generation capabilities across different graph-related tasks.
Building on this flexibility, we can further replace our adopted LoRA~\cite{hu2021lora} with different parameter-efficient training approaches, such as QLoRA~\cite{dettmers2024qlora}, AdaLoRA~\cite{zhang2023adaptive}, and FourierFT~\cite{gao2024parameter}.
These methods can be helpful to reduce the computational demands while maintaining high performance. 
Additionally, we can explore the incorporation of Reinforcement Learning from Human Feedback (RLHF)~\cite{ouyang2022training} to further enhance the learning process, adjusting model behaviors and improving decision-making in complex graph scenarios.

\noindent\textbf{Datasets Extension.} Leveraging the proposed compact graph description and diverse instruction generation mechanisms, we can effectively include a wider range of attributed graphs via our graph-aware instruction tuning, enriching the diversity of applications across various domains. 
Since our \method introduces a generic graph framework to simultaneously capture the semantic and structural information of attributed graphs across tasks and datasets, it can be applied for tasks across diverse datasets, including biological networks, social networks, and knowledge graphs. 
Furthermore, the zero-shot generalization capabilities of our \method, as evaluated via graph-informed DyVal (demonstrated in Fig.~\ref{fig:dyval-bar} in Section~\ref{exp:main_result}), highlight its potential to perform accurately on more datasets where it was not explicitly trained. This ability to generalize well in zero-shot scenarios underscores the practical utility of \method in navigating and analyzing unexplored graphs.

\section{Experiment}
\label{sec:exp}
In this section, we conduct extensive experiments to validate the effectiveness and efficiency of our proposed \method under diverse conditions, aiming to answer the following three key research questions:

\begin{itemize}[leftmargin=10pt]
  \item\textbf{RQ1: }How does our \method framework perform in comparison to the representative graph-oriented methods for generic graph mining?
  \item\textbf{RQ2: }What are the effects of different model components?
  \item\textbf{RQ3: }Can the compact graph description help the model better understand the graph structure?
\end{itemize}

\begin{table}[t]
\caption{Statistics of the datasets for node classification.}
\centering
    \begin{tabular}{c|c c c c}
    \toprule    
    Dataset         & IMDB           & Freebase          & Cora           & Arxiv         \\
    \hline
    \# Nodes        & 21,420         & 180,098           & 2,708          & 169,343       \\
    \# Edges        & 86,642         & 1,057,688         & 5,429          & 1,166,243     \\
    \# Labeled nodes  & 4,573          & 7,954             & 2,708          & 169,343       \\
    \# Classes      & 5              & 7                 & 7              & 40            \\
    \bottomrule
    \end{tabular}
    \label{tab:stat_nc}
\end{table}
\begin{table}[t]
\caption{Statistics of the datasets for link prediction.}
\centering
    \begin{tabular}{c|c c c c}
    \toprule
    Dataset         & \# Nodes          & \# Edges    & \# Link type   & \# Node type   \\
    \hline
    Yelp            & 82,465            & 30,542,675    & 4             & 4             \\
    MIMIC-III       & 32,267            & 559,290       & 4             & 3             \\
    \bottomrule
    \end{tabular}
    \label{tab:stat_lp}
\end{table}

\begin{table}[t!]
\caption{Statistics of the datasets for graph-to-text.}
\centering
    \begin{tabular}{p{1.1cm}|p{1.0cm} p{0.8cm} p{1.1cm} p{1.1cm} p{1.1cm}}
    \toprule
    Dataset  & \# Graph  & \# Rel  & Avg.\# Nodes     & Avg.\# Triples   & Avg. Length \\
    \hline
    AGENDA   & 40,720      & 7            & 12.37             & 4.48              & 140.36      \\
    WebNLG   & 8,783       & 246          & 5.91              & 2.95              & 13.02        \\
    \bottomrule
    \end{tabular}
\label{tab:stat_gt}
\end{table}

\subsection{Experimental Setup}
\subsubsection{Datasets}
To comprehensively evaluate the effectiveness and efficiency of our \method, 
we utilize two real-world datasets for Heterogeneous Node Classification (i.e., IMDB and Freebase), 
two for Homogeneous Node Classification (i.e., Cora and ogbn-arxiv (abbr. Arxiv)), 
two for Link Prediction (i.e., Yelp and MIMIC-III), and two for Graph-to-Text (i.e., AGENDA and WebNLG). 
Additionally, we apply graph-informed DyVal~\cite{zhu2023dyval} to evaluate \method with two Dynamic Reasoning tasks (i.e., Reachability and Max Sum Path), including four levels with increasing complexity (i.e., D1, D2, D3, and D4). 
The detailed statistics are shown in Table~\ref{tab:stat_nc}, Table~\ref{tab:stat_lp}, and Table~\ref{tab:stat_gt}.
Furthermore, the detailed descriptions of tasks and datasets are provided as follows:

\lowercase\expandafter{\romannumeral1}) \textbf{Heterogeneous Node Classification}: Heterogeneous Node Classification classifies the target node into pre-defined classes, leveraging diverse relationships and attributes within the graph, where the graph consists of multiple types of nodes and edges.
Following~\cite{lv2021we}, we split each dataset for Heterogeneous Node Classification into training/validation/testing sets with a ratio of 24\%/6\%/70\%. 
\begin{itemize}
\item {IMDB}\footnote{https://www.kaggle.com/karrrimba/movie-metadatacsv} is a website about movies and related information, including a subset from the Action, Comedy, Drama, Romance, and Thriller genres. Each labeled movie has one or multiple labels.
\item {Freebase}\footnote{http://www.freebase.com} is a knowledge graph of books, films, music, sports, people, locations, organizations, and businesses. Each labeled book has only one label.
\end{itemize}

\lowercase\expandafter{\romannumeral2}) \textbf{Homogeneous Node Classification}: Different from Heterogeneous Node Classification, Homogeneous Node Classification forecasts the target node's label within the graph containing the same type of nodes and edges.
We employ a 60\%/20\%/20\% training/validation/testing split for Cora, which is consistent with~\cite{he2023harnessing}. For Arxiv, we adopt the public dataset split in~\cite{hu2020open}, which is 54\%/18\%/28\%. 
 \begin{itemize}
\item {Cora}\footnote{http://www.cora.justresearch.com/lander} comprises 2,708 scientific publications classified into one of seven classes–case-based, genetic algorithms, neural networks, probabilistic methods, reinforcement learning, rule learning, and theory, with a citation network consisting of 5,429 links.
\item {Arxiv}\footnote{https://ogb.stanford.edu/docs/leader\_nodeprop/\#ogbn-arxiv} represents the citation network among computer science arXiv papers. Each paper in the dataset is associated with a research category, manually labeled by the authors and arXiv moderators. These research categories are chosen from 40 subject areas. 
\end{itemize}

\lowercase\expandafter{\romannumeral3}) \textbf{Link Prediction}: Link Prediction predicts the likelihood of a future or missing connection between two nodes in a graph. We train all methods using the randomly selected 80\% of links and evaluate them on the remaining 20\% held-out links as suggested in~\cite{yang2020heterogeneous} and~\cite{tan2024walklm}.
\begin{itemize}
\item {Yelp}\footnote{https://www.yelp.com/dataset} is a user-business network collected from Yelp, including four types of nodes, which are businesses, users, locations, and reviews. 
\item {MIMIC-III}\footnote{https://physionet.org/content/mimiciii/1.4} consists of a graph of diseases, patients, and visits, with nodes and relations derived from electronic health records. 
\end{itemize}

\lowercase\expandafter{\romannumeral4}) \textbf{Graph-to-Text}: Graph-to-Text generates a descriptive text based on the information and structure of the graph. 
Following~\cite{li2021few}, we design different few-shot settings with four training instance sizes ranging from 50, 100, 200, to 500.
\begin{itemize}
\item {AGENDA}\footnote{https://github.com/rikdz/GraphWriter/tree/master/data} (Abstract Generation Dataset) is a dataset that links knowledge graphs with paper abstracts from scientific domains. The graphs in AGENDA are automatically extracted from the SciIE information extraction system. Each instance in AGENDA includes the paper's title, entities, graph, and abstract.
\item {WebNLG}\footnote{https://github.com/ThiagoCF05/webnlg} is a crowd-sourced RDF triple-to-text dataset manually crafted by human annotators. The dataset includes graphs from DBpedia with up to seven triples paired with one or more reference texts. We adopt three large domains of data from WebNLG v1.5 for experiments (i.e., Airport, Building, and Food). 
\end{itemize}

\lowercase\expandafter{\romannumeral5}) \textbf{Dynamic Reasoning}: Reachability and Max Sum Path are Dynamic Reasoning tasks generated by graph-informed DyVal~\cite{zhu2023dyval}, which generates test samples dynamically, mitigating the issues of data contamination and static complexity. We adopt a zero-shot setting consistent with DyVal~\cite{zhu2023dyval}, where we produce 500 samples for each task of varying complexity to balance test time and discrepancy.
\begin{itemize}
\item {Reachability} determines if one node can reach another node in the graph. Respond with ``True'' or ``False''.  
\item {Max Sum Path} finds the maximum sum path between two nodes in a graph. The sum value is obtained by summing up the values of nodes in the path. If such a path does not exist, directly answer ``N/A''. 
\end{itemize}

\subsubsection{Evaluation Protocols}
For {Heterogeneous Node Classification} and {Homogeneous Node Classification}, we use two commonly adopted evaluation metrics~\cite{yang2020heterogeneous,lv2021we,tan2024walklm}: Macro-F1 (across all labels) and Micro-F1 (across all nodes).
The F1 score is a metric of the model’s accuracy in binary and multi-class classification tasks, which considers both precision and recall.
For {Link Prediction}, we compute the AUC metric as suggested in~\cite{yang2020heterogeneous,tan2024walklm,tang2024graphgpt}.
AUC measures the area under the ROC curve, indicating the model's ability to distinguish between positive and negative classes across various thresholds.
For {Graph-to-Text}, we report BLEU-4~\cite{papineni2002bleu} as our metric, where BLEU-4 measures the precision of four-word sequences (4-grams) in the generated text compared to the reference text.
Moreover, we evaluate {Reachability} and {Max Sum Path} performance with Accuracy that is consistent with DyVal~\cite{zhu2023dyval}.

\subsubsection{Methods for Comparison}
The following characteristic baseline methods can be classified into three categories: i) GNN-based methods, ii) LLM-based methods, and iii) GNN+LLM-based methods. 

i) \textbf{GNN-based methods}: 
\lh{We follow \cite{tang2024graphgpt,wang2024language,yang2020heterogeneous,mao2023hinormer,lv2021we} in selecting GNN-based methods that are commonly used in homogeneous and heterogeneous network benchmarks.}
    \begin{itemize}
        \item {GraphSAGE}~\cite{hamilton2017inductive} generates node embeddings by sampling and aggregating features from a node's local neighborhood, enabling scalable learning on large graphs.
        \item {GCN}~\cite{kipfsemi} scales linearly in the number of graph edges and learns hidden layer representations that encode both local graph structure and features of nodes.
        \item {GAT}~\cite{velickovic2017graph} utilizes masked self-attention mechanisms to enhance the processing of graph data by addressing limitations in traditional graph convolution methods.
        \item {RevGNN}~\cite{li2021training} captures long-range interactions in graph data and reduces memory complexity with grouped reversible connections, enabling more effective training of deep and wide GNNs.
        \item {HINormer}~\cite{mao2023hinormer} uses graph transformers to learn node representations on heterogeneous information networks by capturing both local structure and heterogeneity. 
        \item {R-GCN}~\cite{schlichtkrull2018modeling} improves traditional GCNs with relation-specific convolutions, enhancing learning from diverse edge types in knowledge graphs.
        \item {HGT}~\cite{hu2020heterogeneous} extends the transformer architecture to handle heterogeneous graphs by incorporating type-specific parameters and an attention mechanism to capture diverse node and edge interactions.
    \end{itemize}

ii) \textbf{LLM-based methods}:
\lh{We choose representative and widely adopted LLM-based methods as baselines, covering both general-purpose and graph-specific LLMs. Note that we include the BART~\cite{lewis2019bart} and T5~\cite{ribeiro2020investigating} series as standard encoder-decoder baselines for Graph-to-Text, consistent with~\cite{li2021few,colas2022gap}.}
    \begin{itemize}
        \item {Baichuan2-7B-Base}~\cite{yang2023baichuan} is an open-source, bilingual language model developed by Baichuan Inc., trained on 2.6 trillion tokens with 7 billion parameters. 

        \item {Qwen2-7B-Instruct}~\cite{yang2024qwen2} is an instruction-tuned 7 billion parameter model, designed to excel in tasks like language understanding, generation, and more, with support for processing up to 131,072 tokens in context. 
        
        \item {LLaMA1-7B}~\cite{touvron2023llama} is an open-source large language model developed by Meta. It is designed for natural language understanding and generation tasks, featuring 7 billion parameters for efficient and powerful text processing.
        
        \item {LLaMA2-7B}~\cite{touvron2023llama2} is an improved version of LLaMA1, featuring 7 billion parameters with enhancements in data, training techniques, and model performance.
        
        \item {LLaMA3-8B}~\cite{dubey2024llama}  succeeds LLaMA2, offering improved performance with 8 billion parameters through advancements in architecture, training data, and optimization. 
        
        \item {GPT-3.5}~\cite{ouyang2022training} is a large-scale language model developed by OpenAI, trained on a vast corpus of text with 175 billion parameters, capable of generating human-like text and understanding complex contexts.

        \item {GPT-4}~\cite{achiam2023gpt} builds upon GPT-3.5, providing advanced language generation and understanding capabilities with greater scale and improved performance.

        \item {BART-large}~\cite{lewis2019bart} is a pre-trained language model based on the transformer architecture, featuring both encoder and decoder components, and encompasses 160 million parameters, making it a powerful tool for NLP tasks.

         \item {T5-large}~\cite{ribeiro2020investigating} is a transformer-based pre-trained language model that uses a unified architecture for encoding and decoding, consisting of a 768-layer deep network with 11 billion parameters, effectively handling text-based tasks.

    \end{itemize}
    
iii) \textbf{GNN+LLM-based methods}: 
\lh{We compare recent LLM-enhanced or instruction-tuned graph learning paradigms, as summarized in Table~\ref{tab:comparison}.
For methods such as GPT4Graph~\cite{guo2023gpt4graph} and NLGraph~\cite{wang2024language}, we exclude them as they are general-purpose benchmark frameworks rather than concrete graph models with training pipelines.
LLM-GNN~\cite{chen2023label} and TAPE~\cite{he2023harnessing} rely on GPT-3.5 APIs to generate high-quality annotations and explanations for each dataset. Therefore, we only utilize the publicly released features for the Cora and Arxiv datasets\footnote{https://github.com/CurryTang/LLMGNN}\footnote{https://github.com/XiaoxinHe/TAPE}, and mark ``-'' for the other datasets in Table~\ref{tab:exp-all-nc} and Table~\ref{tab:exp-all-lp}.
Additionally, since GHGRL~\cite{gao2025bootstrapping} is designed with node-centric prompts for node classification, making it unsuitable for link prediction, we report ``-'' in Table~\ref{tab:exp-all-lp}.}
\begin{itemize}
    \item \lh{LLM-GNN~\cite{chen2023label} involves LLMs to generate confidence-aware
annotations for a subset of nodes, which are then used to train GNNs for downstream prediction.}
    \item \lh{{TAPE}~\cite{he2023harnessing} leverages LLMs' explanations to generate informative node features for text-attributed graphs, boosting the performance of various GNNs.}
    \item \lh{{OFA}~\cite{liu2023one} describes diverse text-attributed graphs using human-understandable prompts and encodes them into a unified embedding space via LLMs, thereby guiding the training of a single GNN model.}
    \item \lh{{All-in-One}~\cite{sun2023all}  converts different-level tasks to the graph-level task with the unified graph and language prompts for improving the multi-task performance of GNNs.}
    \item \lh{{GHGRL}~\cite{gao2025bootstrapping} employs LLMs to automatically summarize and classify heterogeneous data, and apply a specialized GNN for task-specific learning.}
    \item {GraphGPT}~\cite{tang2024graphgpt} integrates LLMs with graph knowledge using a graph structural instruction tuning paradigm, enhancing understanding through text-graph grounding and step-by-step reasoning. 
    \item {HiGPT}~\cite{tang2024higpt} aligns LLMs with heterogeneous graph knowledge using instruction tuning, a specialized graph tokenizer, and a mixture-of-thought augmentation to improve understanding and tackle data sparsity. 
    \item \lh{{InstructGLM}~\cite{ye2023natural} tunes LLMs with highly scalable graph prompts that combine natural language instructions and node features from trained GNNs.} 
    \end{itemize}

\subsubsection{Implementation Details}

For our \method, we utilize LLaMA-Factory~\cite{zheng2024llamafactory} to train a unified framework using a mixture of instruction packages across various tasks and datasets.
By default, we choose LLaMA3-8B\footnote{https://ai.meta.com/blog/meta-llama-3}~\cite{dubey2024llama} as the foundation model for fine-tuning, and we perform parameter-efficient learning via LoRA~\cite{hu2021lora} with $r = 32$ and $\alpha = 64$.
The learning rate is set to $5e^{-5}$ and the maximum input length of LLM is set to 1200.
The training process is carried out for two epochs. 
For GNN-based methods, we train and evaluate baselines based on CogDL~\cite{cen2023cogdl}, HGB~\cite{lv2021we}, or HNE~\cite{yang2020heterogeneous}.
For LLM-based methods, we load the checkpoint of LLM from HuggingFace\footnote{https://huggingface.co} 
or call official API from OpenAI\footnote{https://platform.openai.com} for evaluation.
\lh{Additionally, for a fair comparison, we select GNN+LLM-based methods with public checkpoints or tuning details from their original papers. 
All LLM-based methods, GNN+LLM-based methods, and \method, are evaluated using the same test instructions.}
The hyperparameters of baselines are chosen carefully based on either grid search or their official source codes, and the learning rate is searched in $[1e^{-5}, 1e^{-2}]$.
All experiments are conducted using only one NVIDIA GTX 3090 Ti GPU.

Notably, given the GPU constraints during the training phase, we leverage HiGPT~\cite{tang2024higpt} checkpoint from Vicuna-7B-v1.5, which was fine-tuned with 60-shot instruction data on the IMDB\footnote{https://huggingface.co/Jiabin99/HiGPT}, GraphGPT~\cite{tang2024graphgpt} checkpoint tuned with Arxiv-PubMed-mix-NC-LP instruction data\footnote{https://huggingface.co/Jiabin99/GraphGPT-7B-mix-all}, \lh{and {InstructGLM}~\cite{ye2023natural} public checkpoint\footnote{https://github.com/agiresearch/InstructGLM}} for conducting respective experiments.
The full code for this work is available\footnote{https://github.com/Melinda315/MuseGraph}.

\begin{figure}
    \centering
    \includegraphics[width=1\linewidth]{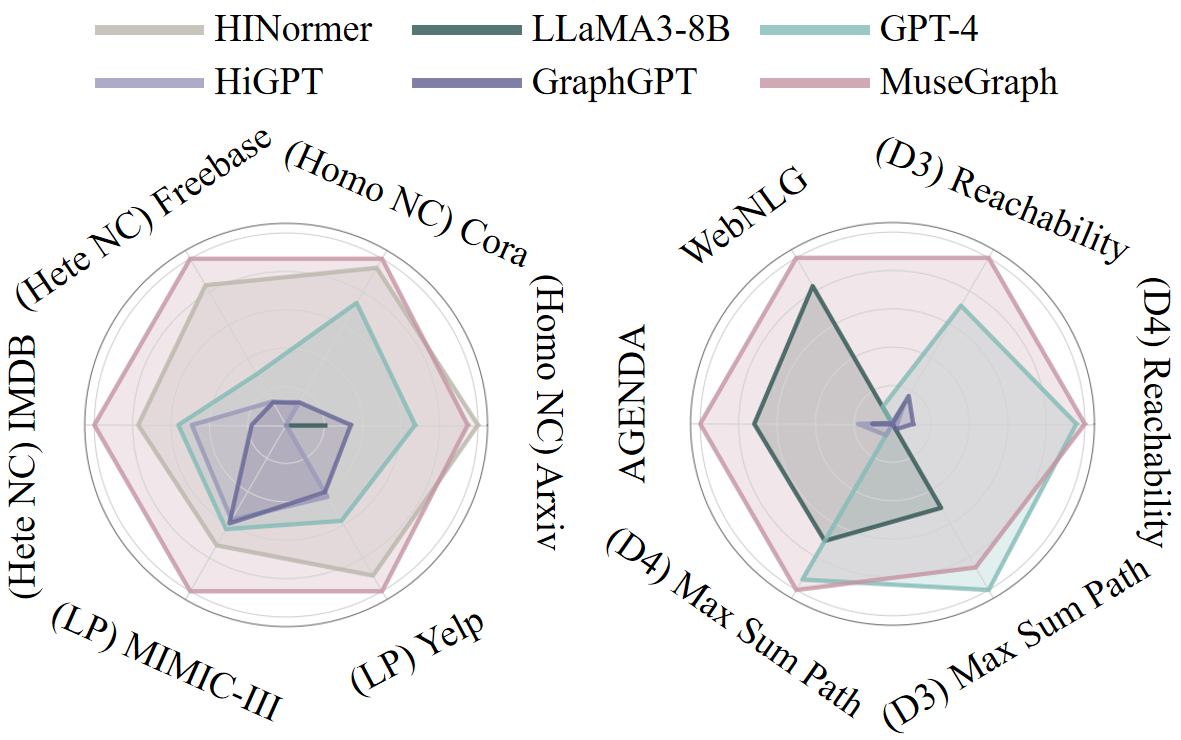}
     \caption{{Comprehensive performance of different models on various tasks and datasets.} The radar charts compare the performance of six models (i.e., HiNormer, HiGPT, LLaMA3-8B, GraphGPT, GPT-4, and \method) across multiple tasks. Micro-F1 scores are shown for Node Classification on IMDB (Heterogeneous) and Cora (Homogeneous). Link Prediction on Yelp uses AUC scores. For limited-data Graph-to-Text tasks on AGENDA and WebNLG, we report BLEU-4 scores. Additionally, accuracy measures for Dynamic Reasoning tasks like Reachability and Max Sum Path (D3 to D4 complexities) from the DyVal benchmark are included. Results are normalized for clarity.
     }
    \label{fig:radar}
\end{figure}

 \begin{table*}
  \caption{Experimental results on four benchmark datasets for Heterogeneous Node Classification and Homogeneous Node Classification. The best performances are highlighted in \textbf{boldface} and the second runners are \underline{underlined}.}
  \centering
  \begin{tabular}{l|cc cc cc cc}
  \toprule
     Method  & Micro-F1 & Macro-F1 & Micro-F1 & Macro-F1 & Micro-F1 & Macro-F1 & Micro-F1 & Macro-F1 \\
     \hline
     
     Dataset & \multicolumn{2}{c}{(Hete NC) IMDB} & \multicolumn{2}{c}{(Hete NC) Freebase} & \multicolumn{2}{c}{(Homo NC) Cora} & \multicolumn{2}{c}{(Homo NC) Arxiv} \\
     \hline
     GraphSAGE                    & 62.06	            & 52.28             & 58.97	            & 27.59              & 85.98	            & 84.57               & 71.69	            & 50.82              \\
     GCN                          & 64.23	            & 58.43             & 59.86	            & 29.35              & 82.87	            & 81.29               & 71.98	            & 51.26              \\
     GAT                          & 64.28	            & 58.81             & 65.83	            & 40.40              & 82.50	            & 81.42               & 72.13	            & \underline{52.67}  \\
     RevGNN                       & 65.03	            & 59.90             & 56.02	            & 20.85              & 85.79	            & 83.14               & \underline{72.76}   & 50.38              \\
     HINormer                     & 67.63               	& 64.29             & 66.51         	& 48.87              & 82.84                & 81.28               & 71.01               & 51.81              \\
     \lh{R-GCN}                   & \lh{62.98}           & \lh{58.90}        & \lh{58.57}        & \lh{47.03}         & \lh{80.63}           & \lh{79.14}          & \lh{71.04}          & \lh{50.79}         \\
     \lh{HGT}                     & \lh{66.95}           & \lh{62.71}        & \lh{60.46}        & \lh{29.33}         & \lh{81.52}           & \lh{80.76}          & \lh{71.80}          & \lh{51.43}         \\

     \hline
     Baichuan2-7B-Base            & 40.57               & 39.15             & 3.60              & 1.62               & 8.78                 & 5.68                & 1.20                & 0.82              \\   
     Qwen2-7B-Instruct            & 64.83               & 61.39             & 20.85             & 13.47              & 58.34                & 48.23               & 40.42               & 18.74              \\  
     LLaMA1-7B                    & 22.57	           & 16.76             & 4.22	           & 1.72               & 12.92	               & 5.93                & 1.35	               & 1.17               \\
     LLaMA2-7B                    & 31.03	           & 27.46             & 5.29	           & 1.78               & 14.02	               & 6.08                & 1.82	               & 1.21               \\
     LLaMA3-8B                    & 37.51	           & 34.86             & 5.98	           & 3.76               & 14.76	               & 8.49                & 23.03	           & 11.09              \\
     GPT-3.5                      & 55.96               & 54.14             & 30.61             & 25.34              & 65.31                & 55.34               & 43.17	           & 32.82              \\
     GPT-4                        & 59.47               & 59.18             & 28.02             & 24.19              & 67.71                & 56.41               & 51.29               & 43.44             \\
     \hline
     \lh{LLM-GNN}       & \lh{-}           & \lh{-}        & \lh{-}        & \lh{-}         & \lh{75.61}           & \lh{73.95}          & \lh{66.05}          & \lh{45.33}             \\
     TAPE (GCN)                   & -	                & -                 & -	                & -                  & \textbf{88.19}	    & \textbf{87.16}      & \textbf{74.36}	    & \textbf{55.74}     \\
     \lh{OFA}                     & \lh{64.81}           & \lh{57.91}        & \lh{56.58}        & \lh{34.75}         & \lh{71.68}           & \lh{70.21}          & \lh{69.31}          & \lh{50.70}             \\
     \lh{All-in-One}              & \lh{60.75}           & \lh{55.17}        & \lh{50.48}        & \lh{21.04}         & \lh{68.45}           & \lh{64.18}          & \lh{65.57}          & \lh{47.06}             \\
     \lh{GHGRL}                   & \lh{65.40}           & \lh{59.24}        & \lh{55.96}        & \lh{26.15}         & \lh{83.96}           & \lh{80.97}          & \lh{63.10}          & \lh{42.05}             \\
     HiGPT                        & 56.72	            & 53.77             & 16.17	            & 7.02               & 23.86	            & 14.17               & 10.76	            & 6.25               \\
     GraphGPT                     & 44.52	            & 43.70             & 15.80	            & 6.14               & 24.57	            & 15.26               & 31.08	            & 17.12               \\
     \lh{InstructGLM}             & \lh{13.59}           & \lh{9.41}         & \lh{5.65}         & \lh{2.38}          & \lh{15.62}           & \lh{10.88}          & \lh{42.60}          & \lh{28.43}             \\
     \hline
     \method (Qwen2-7B-Instruct)  & \underline{75.77}   & \underline{73.08} & \underline{76.40} & \underline{58.18}  & 84.35               & 80.71               & 64.88               & 45.41               \\
     \method (LLaMA1-7B)          & 75.73               & 72.51             & 75.42             & 51.36              & 81.29                & 77.50               & 58.82               & 35.54              \\
     \method (LLaMA2-7B)          & 74.65               & 72.44             & 73.68             & 48.44              & 84.06                & 81.25               & 65.45               & 43.94              \\
     \method (LLaMA3-8B)          & \textbf{76.57}      & \textbf{73.78}    & \textbf{78.01}	   & \textbf{59.62}     & \underline{86.83}	   & \underline{84.74}   & 67.80               & 47.71              \\
 \bottomrule
 \end{tabular}
\label{tab:exp-all-nc}
\vspace{-0.2cm}
\end{table*}

\begin{table}[t]
  \caption{AUC results on two benchmark datasets for Link Prediction. The best performances are highlighted in \textbf{boldface} and the second runners are \underline{underlined}.}
  \centering
  \begin{tabular}{l|c c }
  \toprule
     Dataset                            & Yelp               & MIMIC-III          \\
     \hline
     \lh{GraphSAGE}                     & \lh{68.85}	        & \lh{53.40} \\
     \lh{GCN}                           & \lh{69.94}	        & \lh{53.27} \\
     GAT                                & 70.38              & 54.46             \\
     \lh{RevGNN}                        & \lh{68.97}	        & \lh{56.79} \\
     HINormer                           & 75.03              & 57.82             \\
     R-GCN                              & 72.17	            & 57.31             \\
     HGT                                & \underline{79.02}	& 64.01             \\
     
     \hline
     Baichuan2-7B-Base                  & 14.50              & 12.55             \\
     Qwen2-7B-Instruct                  & 11.18              & 18.31             \\
     LLaMA1-7B                          & 10.78              & 19.39             \\
     LLaMA2-7B                          & 22.14	            & 25.08             \\
     LLaMA3-8B                          & 26.64              & 26.05             \\
     GPT-3.5                            & 50.82              & 56.83             \\
     GPT-4                              & 57.44              & 53.53             \\
     \hline
     \lh{LLM-GNN}       & \lh{-}           & \lh{-} \\
     \lh{TAPE (GCN)}  & \lh{-} & \lh{-}\\
     \lh{OFA}                           & \lh{71.56}	        & \lh{59.33} \\
     \lh{All-in-One}                    & \lh{70.83}	        & \lh{56.20} \\
     \lh{GHGRL}  & \lh{-} & \lh{-}\\
     HiGPT                              & 49.73              & 51.04             \\
     GraphGPT                           & 48.24              & 52.01             \\
     \lh{InstructGLM}                   & \lh{49.38}	        & \lh{34.35} \\
     \hline
     \method (Qwen2-7B-Instruct)        & 70.87              & \textbf{80.21}    \\
     \method (LLaMA1-7B)                & 76.52              & 65.56             \\
     \method (LLaMA2-7B)                & 70.09              & 67.84             \\
     \method (LLaMA3-8B)                & \textbf{80.07}     & \underline{69.92} \\
 \bottomrule
 \end{tabular}
\label{tab:exp-all-lp}
\vspace{-0.15cm}
\end{table}

\begin{table*}
  \caption{BLEU-4 scores for the Graph-to-Text task on AGENDA and WebNLG datasets. GPT-4, HiGPT, and GraphGPT results reflect performances without fine-tuning. In contrast, models like BART-large, T5-large, Qwen2-7B-Instruct, LLaMA3-8B, and \method demonstrate varied performances across different training instance sizes ranging from 50 to 500. The best performances are highlighted in \textbf{boldface} and the second runners are \underline{underlined}.}
  \centering
  \begin{tabular}{l|cccc  cccc}
  \toprule
  \footnotesize
        Dataset                         & \multicolumn{4}{c}{AGENDA}               & \multicolumn{4}{c}{WebNLG}                                     \\
        \hline
        \# Instances                    & 50    & 100   & 200   & 500              & 50      & 100    & 200    & 500                                \\
        \hline
        GPT-4                           & \multicolumn{4}{c}{7.87}                 & \multicolumn{4}{c}{5.43}                                       \\
        HiGPT                           & \multicolumn{4}{c}{8.49}                 & \multicolumn{4}{c}{2.66}                                       \\
        GraphGPT                        & \multicolumn{4}{c}{8.23}                 & \multicolumn{4}{c}{2.45}    \\
        \lh{InstructGLM}      & \multicolumn{4}{c}{\lh{5.97}}     	& \multicolumn{4}{c}{\lh{5.68}} 
        \\
        \hline
        BART-large                      & 9.91              & 10.58             & 11.99             & 12.49                   & 26.59              & 28.53              & 30.62              & 33.15                 \\
        T5-large                        & 1.82              & 5.59              & 7.75              & 9.55                    & 21.27              & 23.57              & 26.48              & 31.64                 \\
        Qwen2-7B-Instruct               & 7.95              & 8.28              & 9.90              & 12.41                   & 27.24              & 32.29              & \underline{34.90}  & 36.02                \\
        LLaMA3-8B                       & 10.35             & 10.44             & \underline{13.07}             & \underline{13.36}       & 26.53              & \underline{33.43}  & 33.55              & 36.54                \\
        \hline
        \method (Qwen2-7B-Instruct)     & \underline{10.77} & \underline{11.34} & {12.78} & 12.86                   & \underline{31.15}  & 32.35              & \textbf{35.89}     & \textbf{36.87}       \\
        \method (LLaMA3-8B)             & \textbf{11.32}    & \textbf{13.11}    & \textbf{13.28}    & \textbf{14.64}          & \textbf{31.46}    & \textbf{33.97}     & 34.25              & \underline{36.83}     \\
    \bottomrule
  \end{tabular}
\label{tab:exp-few-shot-gt}
\end{table*}

\subsection{Main Results Across Different Tasks (RQ1)}
\label{exp:main_result}
In this subsection, we provide a comprehensive performance analysis of our proposed \method framework across various graph tasks, evaluating its generation and adaptability capabilities in general, few-shot, and zero-shot settings compared with state-of-the-art baselines.

Overall, our proposed \method consistently demonstrates superior performance, highlighting its robust understanding of graph data combined with strong language generation capabilities (shown in Fig.~\ref{fig:radar}).
In Node Classification and Link Prediction tasks across six datasets, \method achieves an average ranking of 2.17. 
In few-shot Graph-to-Text tasks and Dynamic Reasoning tasks in the zero-shot setting, \method secures the first and second rankings, respectively. 
Specifically, despite GPT-4's strength in handling attribute-rich datasets and outperforming our \method in most Dynamic Reasoning tasks, it often struggles with adaptability across unfamiliar tasks and settings. 
The ability of \method to match GPT-4 in complex Dynamic Reasoning tasks with significantly fewer trainable 13,631,488 parameters and training data highlights not just resource efficiency but also a practical solution for real-world graph applications requiring high adaptability and lower computational demands.

Similarly, domain-specific models such as HiGPT and GraphGPT perform commendably within their respective training contexts (e.g., IMDB) but falter when faced with scenarios outside these predefined datasets. HINormer, while effective at managing complex relationships in graph-specific tasks, lacks the versatility needed for broader applications (e.g., Graph-to-Text and Dynamic Reasoning tasks). In contrast, LLaMA3-8B excels in Graph-to-Text due to its extensive linguistic pre-training but, like others, it encounters limitations when stepping beyond its core competencies. 
\method, however, maintains robust performance across a variety of tasks, effectively integrating and generalizing graph data and language generation abilities where other models show constraints. 

\begin{figure}[t]
    \centering
\includegraphics[width=0.8\linewidth]{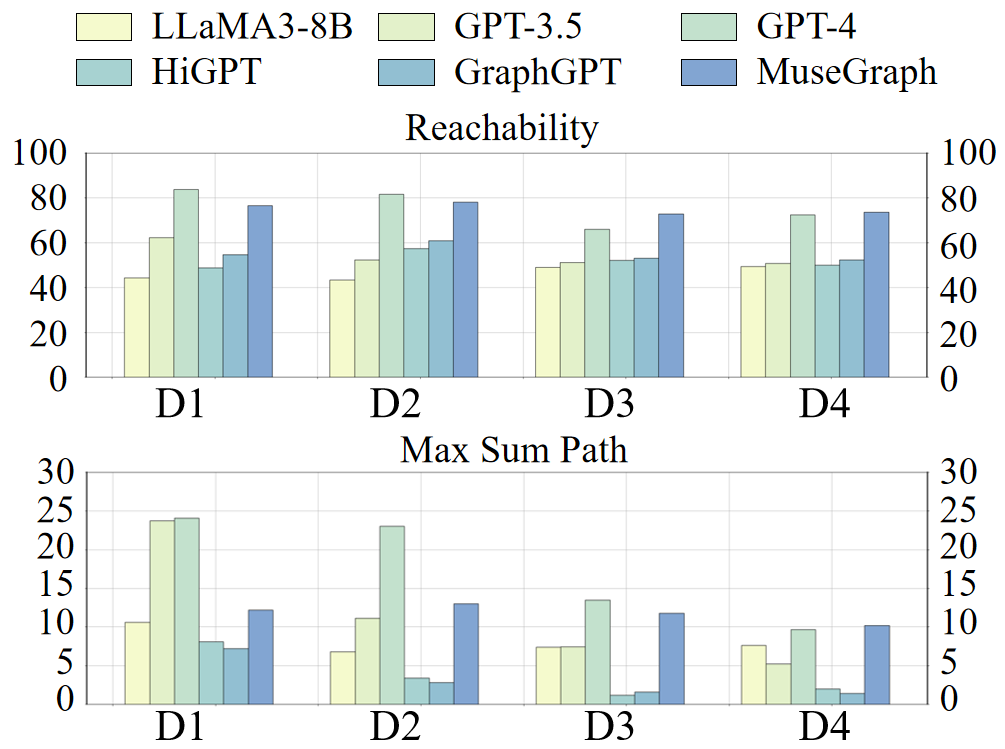}
     \caption{Accuracy Results for the Reachability and Max Sum Path tasks on the graph with varying difficulty levels (i.e., D1 to D4), where ``D'' represents different degrees of complexity.}
    \label{fig:dyval-bar}
\end{figure}

\subsubsection{Compared with GNN-based Methods}

As shown in Table~\ref{tab:exp-all-nc} and Table~\ref{tab:exp-all-lp}, our \method outperforms GNN-based methods across most tasks and datasets, demonstrating its accurate comprehension of graph data.
\lh{Notably, \method achieves significant performance gains in Heterogeneous Node Classification on IMDB and Freebase with an average of 13.99\% and 19.64\%. 
Moreover, \method consistently outperforms all GNN-based baselines in Link Prediction with an average performance improvement of 17.54\%.}

\lh{Specifically, while approaches like HINormer excel in Heterogeneous Node Classification, GraphSAGE and GAT in Homogeneous settings, and HGT in Link Prediction, they generally struggle with Graph-to-Text and Dynamic Reasoning tasks due to their limited adaptability and lack of linguistic features.}
In contrast, \method that can be easily tuned on different foundation models with a small number of corpora demonstrates a comprehensive understanding of graph data coupled with language generation capabilities.
\lh{The overall performance across a variety of tasks and datasets makes \method a more practical solution than traditional GNN-based methods like HINormer and GAT.}

\subsubsection{Compared with LLM-based Methods}
Generally, \method surpasses all LLM-based methods with significant improvements on both general graph tasks and few-shot ones, highlighting \method's superior capabilities in generic graph mining (shown in Table~\ref{tab:exp-all-nc}, Table~\ref{tab:exp-all-lp}, and Table~\ref{tab:exp-few-shot-gt}).
Note that, by effective graph-aware tuning, our \method outperforms its foundation model LLaMA3-8B, ranging from 0.79\% in BLEU-4 under 500 training instances on WebNLG to 1485.64\% in Macro-F1 on Freebase.

In detail, GPT-4, known for its extensive parameter set and strong generalization ability, excels in Dynamic Reasoning tasks and attribute-rich datasets (as depicted in Fig.~\ref{fig:dyval-bar} and Table~\ref{tab:exp-all-nc}).
However, its large scale and closed-source nature often preclude fine-tuning, which limits its adaptability to complex graph structures and unfamiliar tasks.
For example, when accessed via its official API without fine-tuning, GPT-4 produces verbose and inaccurate text in few-shot Graph-to-Text tasks, resulting in suboptimal BLEU-4 scores due to hallucinatory content.
By comparison, open-source LLMs (e.g., Qwen2-7B-Instruct and LLaMA3-8B) that can be slightly fine-tuned demonstrate improved performance and reduced training costs.
For instance, BART-large and LLaMA3-8B, tailored with a specific corpus of Graph-to-Text, gain average improvements over GPT-4 with 245.11\% and 274.38\%, respectively.
However, without our proposed diverse instruction generation mechanism, these LLMs still encounter challenges in fully comprehending complex graph structures, reflecting inherent limitations in their adaptability to diverse graph data.

Note that, \method employs a graph-aware instruction tuning mechanism that utilizes fewer but more diverse CoT-based instruction packages and fewer trainable parameters. This approach not only enhances the precision of graph-oriented downstream tasks with a single NVIDIA GTX 3090 Ti GPU but also boosts generation capabilities, enabling \method to effectively match GPT-4 in complex Dynamic Reasoning tasks with improved efficiency and adaptability.

\subsubsection{Compared with GNN+LLM-based Methods}
\lh{Overall, our proposed \method exceeds the majority of GNN+LLM-based methods across various tasks and datasets, demonstrating its robust understanding of diverse graph data with powerful language generation capabilities (as depicted in Table~\ref{tab:exp-all-nc}, Table~\ref{tab:exp-all-lp}, Table~\ref{tab:exp-few-shot-gt}, and Fig.~\ref{fig:dyval-bar}).
For Heterogeneous Node Classification, the performance improvements
range from 17.08\% in Micro-F1 on IMDB to 71.57\% in Macro-F1 on Freebase. 
Additionally, \method significantly outperforms all GNN+LLM-based methods in both Graph-to-Text and Dynamic Reasoning tasks, benefiting from its superior language generation and reasoning capabilities.
}

\lh{Some methods (e.g., LLM-GNN, TAPE, OFA, and GHGRL) train GNNs as predictors with LLM-enhanced features, combining the interpretative strengths of LLMs with GNNs to improve text-attributed graph representation learning.
Particularly, OFA unifies different graph data by describing nodes and edges using natural language and converts them into a shared embedding space via LLMs.
TAPE achieves top performance on node classification datasets like Cora and Arxiv by encoding LLMs' explanations.
However, it relies on manually customized prompts and GPT-3.5 APIs to generate high-quality explanations for each dataset, which incurs high training costs and restricts its adaptability across diverse tasks and datasets.
Moreover, these methods do not apply to Graph-to-Text and Dynamic Reasoning tasks due to their lack of language understanding and generation abilities. 
This further highlights the adaptability and flexibility of \method beyond traditional graph prediction tasks.}

\begin{figure}
    \centering
    \includegraphics[width=0.8\linewidth]{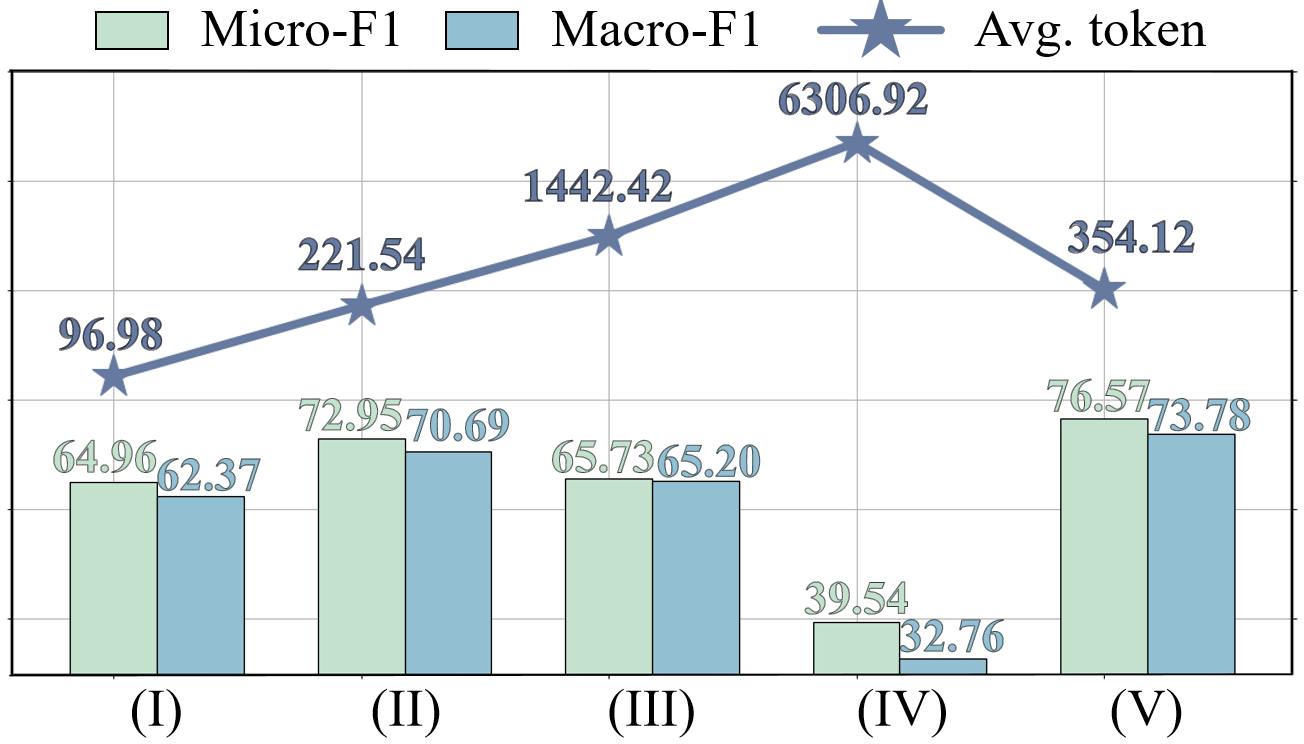}
     \caption{Comparison of results for Heterogeneous Node Classification on IMDB across different graph descriptions as input.
     (I) through (IV) denote varying levels of neighbor inclusion: (I) utilizes central node attributes; (II) integrates one-hop neighbors; (III) includes up to two-hop neighbors; and (IV) encompasses three-hop neighbors. (V) employs a proposed method combining one-hop neighbors and random walks. Micro-F1 and Macro-F1 scores are shown alongside the average number of tokens used on IMDB.
     }
    \label{fig:abl_chart_compact_graph}
\end{figure}

\lh{Another category of methods, like HiGPT, GraphGPT, and InstructGLM, utilizes LLMs to directly perform graph tasks. These methods typically depend on task/dataset-specific GNNs for graph encoding and integrate node representations into the input instructions, which leads to significant computational overhead and limited cross-dataset adaptability.}
By contrast, as illustrated in Table~\ref{tab:exp-few-shot-gt}, \method achieves substantial performance gains by efficiently fine-tuning with even a limited number of examples. This ability highlights its effectiveness in adapting to different Graph-to-Text tasks and securing significant improvements in model performance.
Note that HiGPT demonstrates flexibility in zero-shot learning across heterogeneous graph datasets, like when trained on IMDB and tested on DBLP~\cite{tang2024higpt}. However, it shows significant performance drops on homogeneous datasets such as Cora (shown in Table~\ref{tab:exp-all-nc}), which highlights its limited adaptability to different graph structures. Additionally, HiGPT struggles with Link Prediction tasks due to a lack of task-specific training corpus, often yielding incorrect ``Yes'' responses. 

These issues underline the critical need for GNN+LLM-based methods to incorporate task-specific training, boost computational efficiency, and achieve generalization across diverse graph structures. In contrast, \method introduces a compact graph description without specific GNNs to generate fewer diverse and high-quality instruction data under limited language tokens for fine-tuning.
Our proposed approach effectively alleviates the content truncation problem while facilitating efficiency and adaptability across various graph tasks and datasets.

\subsection{Ablation Studies (RQ2)}
To demonstrate the impact of various model components, we conduct a series of ablation studies, including (1) Variations in Foundation Models, (2) Variations in Compact Graph Descriptions, (3) Variations in CoT-based Instruction Packages, and (4) Variations in Dynamic Instruction Package Allocation Strategies. We have the following observations.

\begin{table}
  \caption{Ablation study results for Heterogeneous Node Classification (NC) on IMDB under different settings of fine-tuning. ``NC pkg w/o. CoT'' excludes Chain of Thought (CoT)-based instructions, while ``NC pkg'', ``NC pkg + LP pkg (1:1)'', and ``NC pkg + LP pkg (3:1)'' include them. Notably, ``NC pkg + LP pkg (1:1)'' and ``NC pkg + LP pkg (3:1)'' integrate Link Prediction (LP) tasks specifically tailored for the IMDB dataset. The allocation ratio 3:1 denotes that for every four CoT-based instruction packages used, three are from the NC task and one is from the LP task. In contrast, our \method (LLaMA3-8B) employs CoT-based instructions driven by multiple datasets and tasks, not limited to IMDB. Performance is measured in Micro-F1 and Macro-F1 scores, with the average token count on IMDB provided.
  }
  \centering
  \begin{tabular}{l|ccc}
  \toprule
     Setting             & Micro-F1              & Macro-F1         & Avg. token  \\
     \hline                
     NC pkg w/o. CoT              & 71.23                 & 70.62            & 317.87        \\
     NC pkg               & 73.03                 & 71.45            & 339.73        \\
     \hline
     NC pkg + LP pkg (1:1)         & 72.81                 & 71.07            & 346.09        \\
     NC pkg + LP pkg (3:1)         & 74.86                 & 72.13            & 354.12        \\
     \method (LLaMA3-8B)             & 76.57                 & 73.78            & 354.12        \\
\bottomrule
\end{tabular}
\label{tab:ablation-across-task-dataset}
\end{table}

\begin{figure*}[ht]
    \centering
\includegraphics[width=1\linewidth]{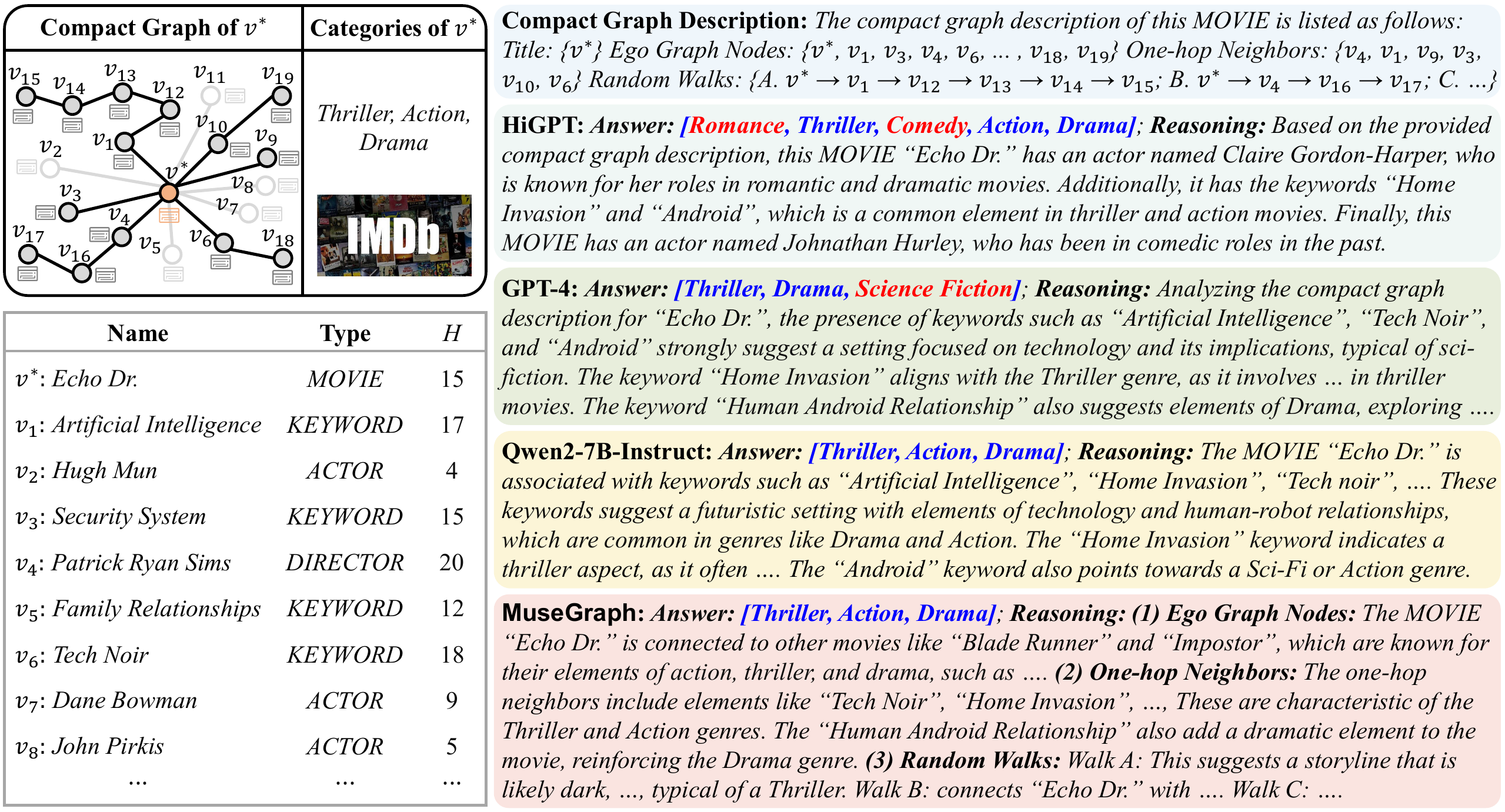}
    \caption{Heterogeneous Node Classification Results on IMDB from HiGPT, GPT-4, Qwen2-7B-Instruct, and \method, all utilizing our compact graph description as input. These models, selected for their generation and reasoning capabilities of LLMs, are compared based on their ability to analyze graphs. Best viewed in color.}
    \label{fig:case_studies}
\end{figure*}

\subsubsection{Variations in Foundation Models}
\label{exp:foundation_model}
We explore several foundation models, including Qwen2-7B-Instruct, LLaMA1-7B, LLaMA2-7B, and LLaMA3-8B, across various graph-related tasks and datasets.

According to Table~\ref{tab:exp-all-nc}, Table~\ref{tab:exp-all-lp}, and Table~\ref{tab:exp-few-shot-gt}, \method (LLaMA3-8B) outperforms other variations of \method across most tasks and datasets, with performance improvements ranging from 0.96\% in Macro-F1 on IMDB to 15.61\% in BLEU-4 under 100 training instances on AGENDA.
This indicates that a foundation model with larger parameters and extensive training can achieve better and more stable results with our graph-aware instruction tuning.
Moreover, \method (Qwen2-7B-Instruct) also achieves remarkable results on MIMIC-III and WebNLG, underscoring the adaptability of our framework to harness the distinct capabilities of various LLMs, enhancing their robust understanding of graph data.
Through the above results, \method proves to be cost-effective, by utilizing fine-tuning of diverse LLMs rather than relying on extensive new data and training from scratch.

\subsubsection{Variations in Compact Graph Descriptions}
\label{sec:ablation_compact_graph}

We assess various compact graph descriptions on Heterogeneous Node Classification on the IMDB dataset, exploring models with different levels of neighborhood and walk integration. 
The simplest model (I) uses only central node attributes.
Models (II) through (IV) gradually integrate more extensive neighborhood attributes, enhancing the model's context-awareness.
Our advanced model (V) uniquely merges one-hop neighbors with random walks, striking an optimal balance between richness of information and token efficiency, as shown in Fig.~\ref{fig:abl_chart_compact_graph}.

Specifically, compared with description (I), which only has attribute information of the central node, description (II) integrates all one-hop neighbor attributes to achieve up to 13.34\% improvements in Macro-F1.
However, when extending the neighborhood to two-hop or multi-hop without sampling decreases effectiveness (i.e., (III) and (IV)), the performance significantly deteriorates due to an overload of input tokens.
Conversely, model (V) effectively combines one-hop neighbors with random walks within a controlled token budget by calculating the node energy (cf., Eq.~\ref{eq:h}). This ensures a rich information intake and maintains token efficiency, significantly improving performance on complex graph structures.

\subsubsection{Variations in CoT-based Instruction Packages}

In the ablation study detailed in Table~\ref{tab:ablation-across-task-dataset}, we investigate the impact of different variations in Chain of Thought (CoT)-based instruction packages on the IMDB dataset for the Heterogeneous Node Classification task. 

When CoT-based instructions are incorporated with task-specific instructions (i.e., NC pkg w/o. CoT vs. NC pkg), there is a notable improvement in both Micro-F1 and Macro-F1 scores with only a 21.86 increase in the average number of tokens, indicating enhanced model understanding and reasoning capabilities with minimal impact on computational efficiency. 
Note that appropriately integrating CoT-instruction packages for Link Prediction tasks on IMDB (e.g., NC pkg + LP pkg (3:1)) demonstrates a compounded beneficial effect in Heterogeneous Node Classification, as further analyzed in Section~\ref{sec:ablation_dynamic_allocation strategies}.
This not only further improves the model's adaptability across various graph tasks, but also underscores the synergistic effect of diverse yet balanced CoT-based instruction packages on comprehension capabilities for generic graphs.

\subsubsection{Variations in Dynamic Instruction Package Allocation Strategies}
\label{sec:ablation_dynamic_allocation strategies}
As shown in Table~\ref{tab:ablation-across-task-dataset}, we assess the impact of varying CoT-based instruction package compositions, incorporating instruction packages from multiple tasks and datasets, and applying them to perform graph-aware instruction tuning.

By adjusting the allocation ratios of Link Prediction CoT-based instructions (e.g., a 3:1 ratio compared to a 1:1 ratio) within the NC pkg + LP pkg, we observed performance gains for Heterogeneous Node Classification, with Micro-F1 rising from 72.81 to 74.86. This demonstrates the critical need for carefully balancing instruction packages based on the complexity and requirements of each task, enabling LLMs to accurately understand and adapt to diverse graph-based tasks.
Furthermore, when we diversify the instruction set with various graph tasks and datasets, our \method (LLaMA3-8B) can surpass NC pkg + LP pkg (3:1) with a superior performance improvement of 2.29\% on average.
This strongly indicates that the strategic dynamic instruction package allocation with different task complexities and dataset complexities can enhance model adaptability across different graph tasks and datasets while mitigating the risk of catastrophic forgetting.

\subsection{Case Studies (RQ3)}
To evaluate the effectiveness of compact graph descriptions in enhancing LLMs’ comprehension of graph structures, we provide a Heterogeneous Node Classification scenario on the IMDB dataset. Fig.~\ref{fig:case_studies} contrasts the responses from HiGPT, GPT-4, Qwen2-7B-Instruct, and \method, which all interpret the same compact graph description but exhibit varying analytical abilities due to distinct LLM architectures.

Overall, the utilization of compact graph descriptions facilitates a deeper and more accurate analysis by encapsulating key graph data into a more manageable and interpretable format. 
This not only preserves critical information but also enhances the models' understanding and reasoning capabilities to generate relevant and context-aware responses, allowing models to focus on the pertinent details without the noise commonly associated with extensive graph data. 
As shown in the blue responses of Fig.~\ref{fig:case_studies}, both Qwen2-7B-Instruct and \method benefit from this compact graph description, achieving correct predictions and analyses that reflect the insightful understanding of graph data. 

Note that, HiGPT and GPT-4 perform wrong predictions and provide the additional categories for $v^*$ (shown in the red responses of Fig.~\ref{fig:case_studies}).
HiGPT struggles with the compact graph description due to the absence of specific learned graph tokens for input. 
This limitation causes HiGPT to revert to its existing knowledge base for category analysis, leading to irrelevant and misleading suggestions, such as attributing the category ``Comedy'' to Johnathan Hurley based on his past roles, which does not align with $v^*$'s actual categories.
GPT-4, with its extensive parameters and knowledge, excels in studying the semantic information of keywords presented in $v^*$'s compact graph description to predict the categories.
However, this makes it ignore graph structures and leads to hallucinations, resulting in erroneous predictions like ``Science Fiction''.
Although Qwen2-7B-Instruct shows remarkable performances on IMDB, it is hard to adapt to different tasks and datasets without further fine-tuning (depicted in Table~\ref{tab:exp-all-nc}, Table~\ref{tab:exp-all-lp}, and Table~\ref{tab:exp-few-shot-gt}).

In contrast, \method can accurately understand graph structures via compact graph description. For example, the one-hop neighbor ``Home Invasion'' and the walk ``A'' suggest a strong thriller element of node ``$v^*$''. These elements collectively enable \method to provide a detailed and contextually enriched genre prediction, showcasing its ability to synthesize complex graph structures into coherent and accurate analyses.

\section{Conclusion and Future Work}
In this paper, we introduce \method, an effective and generic approach for graph mining, which can enable the accurate understanding abilities of graph data across various tasks and datasets.
Through the innovative design of compact graph descriptions with adaptive generation procedure, the generation of diverse task-specific Chain-of-Thought (CoT)-based instruction packages, and the implementation of graph-aware instruction tuning, our \method integrates the strengths of Graph Neural Networks (GNNs) and Large Language Models (LLMs) into one single foundation model.
Our comprehensive experimental results demonstrate \method's superior performance against state-of-the-art
baselines in five graph tasks and ten datasets, illustrating its ability not only to improve the precision of graph-related downstream tasks but also to enhance the generation capabilities of LLMs, which is further consolidated with our real case study results.

The primary limitation of our \method lies in the reliance on input graphs with rich semantic information and the careful selection of training datasets. 
Looking ahead, it is interesting to broaden the scope of \method by incorporating it with a wider range of graph types and exploring its applications in more diverse tasks. For instance, applying \method to biological graphs could be intriguing, where integrating domain-specific knowledge and expert feedback is crucial. 

\section*{Acknowledgments}
\noindent This work was supported in part by the National Natural Science Foundation of China under Grants (No.62302098) and Fujian Provincial Natural Science Foundation of China under Grants (2025J01540). Carl Yang was not supported by any fund from China.


\bibliographystyle{IEEEtran}
\bibliography{musegraph}

\vspace*{-3ex}
\begin{IEEEbiography}[{\includegraphics[width=0.96in,height=1.1in,clip,keepaspectratio]
{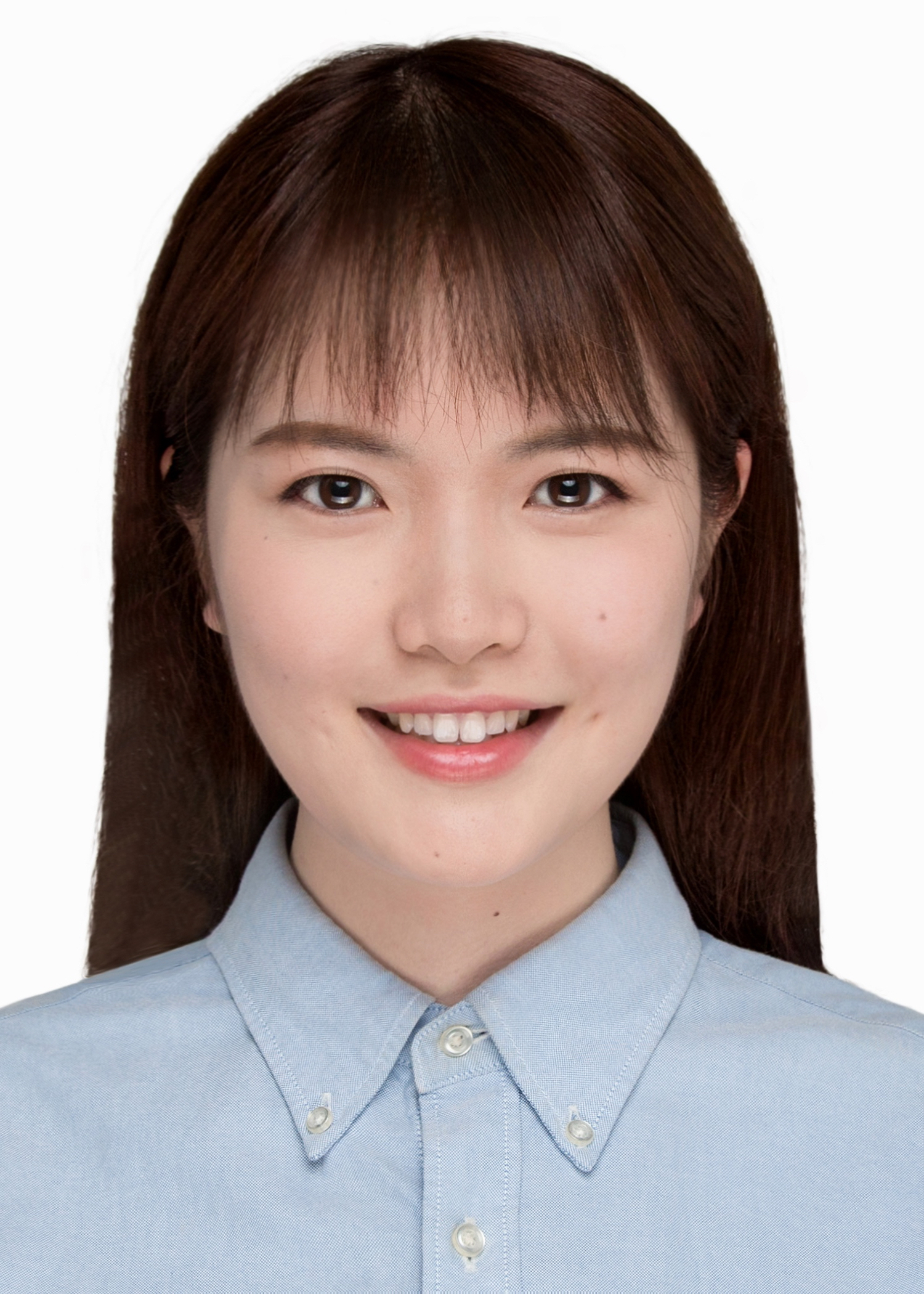}}]{Yanchao Tan} is currently working as an Associate Professor with the College of Computer and Data Science, and Fujian Key Laboratory of Network Computing and Intelligent Information Processing, Fuzhou University, Fuzhou, China. She obtained her Ph.D. degree from the College of Computer Science, Zhejiang University, Hangzhou, China, in 2022. She was a Visiting Scholar at The Chinese University of Hong Kong from 2024 to 2025. Her research interests include data mining, healthcare, recommender systems, and large language models.
\end{IEEEbiography}

\vspace*{-8ex}
\begin{IEEEbiography}[{\includegraphics[width=0.96in,height=1.1in,clip,keepaspectratio]
{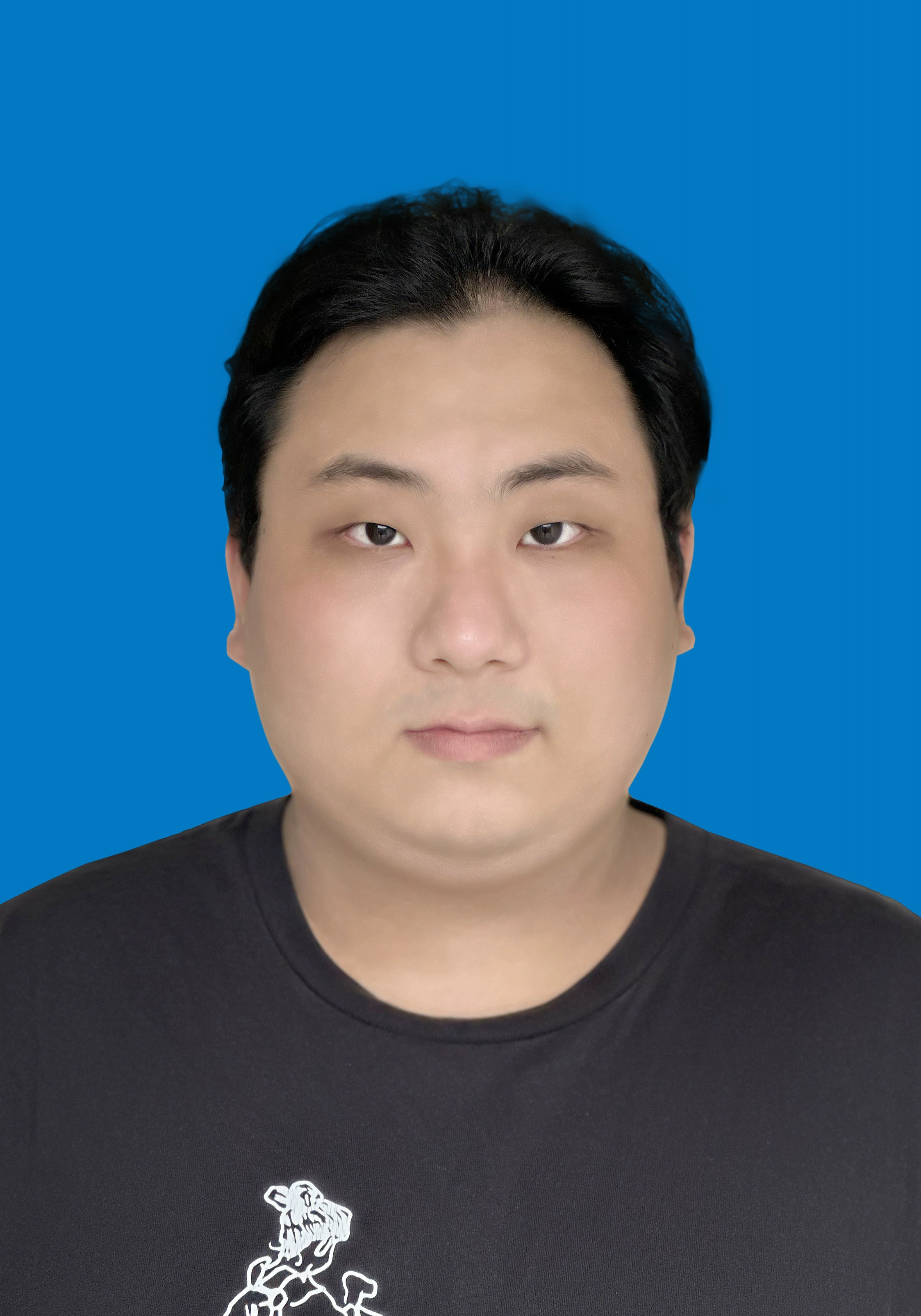}}]{Hang Lv} received a BS degree in computer science from Fuzhou University, China, in 2023. He is currently working toward a Ph.D. degree in computer science and technology at Fuzhou University, China. His research interests include data mining, healthcare, recommender systems, large language models, and graph representation learning.
\end{IEEEbiography}

\vspace*{-8ex}
\begin{IEEEbiography}[{\includegraphics[width=0.96in,height=1.1in,clip,keepaspectratio]
{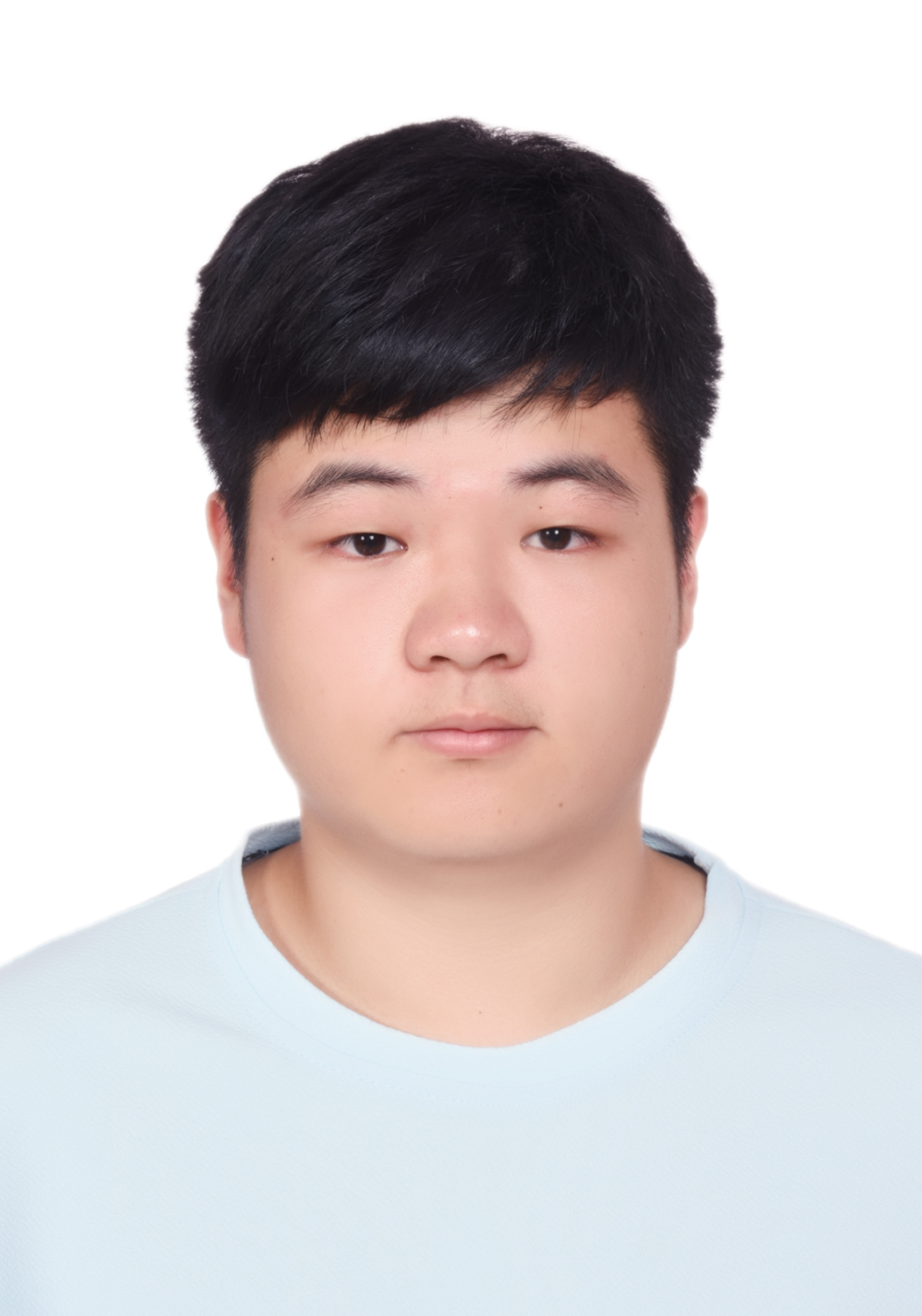}}]{Pengxiang Zhan} received a BS degree in computer science from Fuzhou University, China, in 2023. He is currently pursuing a Master’s degree in computer science and technology at Fuzhou University, China. His research interests include language models and multimodal large models. 
\end{IEEEbiography}

\vspace*{-8ex}
\begin{IEEEbiography}[{\includegraphics[width=0.96in,height=1.1in,clip,keepaspectratio]
{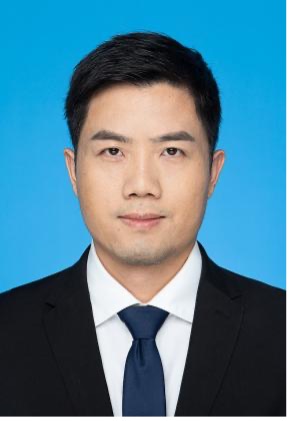}}]{Shiping Wang} is currently working as a Professor with the College of Computer and Data Science, Fuzhou University, Fuzhou, China, and the Director of the Key Laboratory of Intelligent Metro, Fujian Province University, Fuzhou, China. He received his Ph.D. degree from the University of Electronic Science and Technology of China, Chengdu, China in 2014. He was a Visiting Scholar in University of Alberta, Edmonton, Canada from August 2013 to August 2014. He worked as a Research Assistant in National University of Singapore from January 2014 to August 2014, and a Research Fellow in Nanyang Technological University of Singapore from August 2015 to August 2016. He was also a Visiting Researcher in Peking University, Beijing, China from August 2019 to August 2020. His research interests include machine learning, deep learning, feature representation and multi-modal fusion.
\end{IEEEbiography}

\vspace*{-8ex}
\begin{IEEEbiography}[{\includegraphics[width=0.96in,height=1.1in,clip,keepaspectratio]
{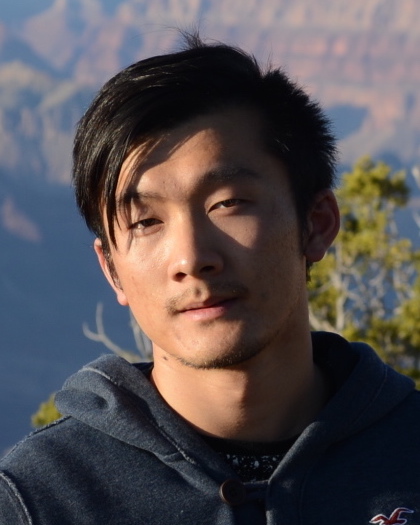}}]{Carl Yang} is an Assistant Professor of Computer Science at Emory University, jointly appointed in the Rollins School of Public Health and Nell Hodgson Woodruff School of Nursing. He received his Ph.D. in Computer Science at University of Illinois, Urbana-Champaign in 2020, and B.Eng. in Computer Science and Engineering at Zhejiang University in 2014. His research interests span data mining, knowledge graphs, multimodality foundation models, and trustworthy AI, with applications in network sciences, neuroscience, biomedicine, and healthcare. Carl's research results have led to 200+ peer-reviewed publications in top venues across AI/ML and medicine/healthcare. He is also a recipient of the SIGKDD Rising Star Award in 2025, NSF CAREER Award in 2025, NIH K25 (Career) Award in 2023, and multiple Best Paper Awards such as of MedInfo 2025, KDD Health Day 2022, ML4H 2022, and ICDM 2020. 
\end{IEEEbiography}

\end{document}